\pdfoutput=1

\documentclass[11pt]{article}

\usepackage[preprint]{acl}

\usepackage{times}
\usepackage{latexsym}

\usepackage[T1]{fontenc}

\usepackage[utf8]{inputenc}

\usepackage{microtype}

\usepackage{inconsolata}

\usepackage{graphicx}
\usepackage{subcaption} 
\usepackage{booktabs}
\usepackage{multirow}
\usepackage{amsmath}
\usepackage{amsfonts}
\usepackage{amssymb}
\usepackage{tcolorbox}
\usepackage{tabularx}

\definecolor{promptaccent}{HTML}{4F7467}
\definecolor{prompttitle}{HTML}{E5EFE9}
\definecolor{promptbody}{HTML}{FAFCFA}

\newtcolorbox{paperprompt}[1]{
    title={#1},
    colback=promptbody,
    colframe=promptaccent!45!white,
    colbacktitle=prompttitle,
    coltitle=promptaccent!65!black,
    boxrule=0.45pt,
    titlerule=0pt,
    arc=1.5pt,
    boxsep=0pt,
    left=7pt, right=7pt,
    top=6pt, bottom=6pt,
    toptitle=4pt, bottomtitle=4pt,
    middle=5pt,
    fonttitle=\small\bfseries,
    fontupper=\footnotesize,
    fontlower=\footnotesize,
    before upper={\raggedright\setlength{\parindent}{0pt}},
    before lower={\raggedright\setlength{\parindent}{0pt}},
    before skip=0pt,
    after skip=0pt
}

\newcommand{\promptcontext}[1]{%
    {\itshape\color{promptaccent!80!black}#1}%
}

\newcommand{\handofflabel}[1]{%
    \par\noindent{\scriptsize\sffamily\color{promptaccent!80!black}#1\par}%
    \nobreak\vspace{1pt}\noindent\ignorespaces
}

\usepackage[capitalise]{cleveref}

\usepackage{todonotes}
\usepackage{soul}
\usepackage[normalem]{ulem}
\usepackage{bm}
\usepackage{dsfont}
\usepackage{mathtools}
\usepackage{pifont}
\newcommand{\hsh}[1]{\textcolor{blue}{#1}}

\newcommand{\yb}[1]{\textcolor{orange}{#1}}

\title{New Model, Old Habits: Behavior Transfer under Model Swapping \\in Multi-Session LLM Agents}

\author{
 \textbf{Youngbin Choi\textsuperscript{1,*}},
 \textbf{Sehyun Hwang\textsuperscript{3,*}},
 \textbf{Suhyeon Jun\textsuperscript{2}},
 \textbf{Seoyeon Lee\textsuperscript{1}},\\
 \textbf{Sunghyun Choi\textsuperscript{1}},
 \textbf{MoonJeong Park\textsuperscript{1}},
 \textbf{Dongwoo Kim\textsuperscript{1,2,\textdagger}}
\\
 \textsuperscript{1}Graduate School of Artificial Intelligence, POSTECH,
\\
 \textsuperscript{2}Department of Computer Science and Engineering, POSTECH, 
\\
 \textsuperscript{3}Meta, \\
 \small{
    \{choi.youngbin, sehyun03, suhyeonjun, seoyeon26, sunghyunchoi, mjeongp, dongwoo.kim\}@postech.ac.kr
 }
}

\begin{document}
\maketitle
\begingroup
\renewcommand\thefootnote{}
\footnotetext{\textsuperscript{*}Equal contribution.}
\footnotetext{\textsuperscript{\textdagger}Corresponding author.}
\endgroup

\begin{abstract}

Multi-session agents can switch between LLMs to balance performance and inference cost.
Such switches require transferring a history for continuity, yet their effects on the recipient remain underexplored.
Through controlled experiments with multi-session tasks, we uncover \textit{behavior transfer}: recipients tend to adopt reasoning and action patterns from the models that generated their histories.
This effect is most evident with raw-history transfer but is substantially reduced by summarization.
Based on these findings, we propose Direction- and Authorship-aware Summarization of History (DASH) that preserves stronger-model behavior while minimizing weak-to-strong transfer via summarization.
Experiments with repeated swaps show that DASH outperforms raw-history transfer and naive summarization baselines.
These results suggest that effective multi-model orchestration requires considering not only which model to use, but also which behavioral patterns to carry forward.



\end{abstract}

\section{Introduction}

Personal agents are evolving from answering individual questions to carrying out real-world tasks over extended periods, guided by users' preferences and constraints~\citep{menghini2026musesparksafety,openai2026dots,spacexai2026grokbot,nousresearchHermesAgent}.
This motivates the study of multi-session tasks, where later decisions depend on earlier outcomes~\citep{he2026benchmarking,li2026agencybench}.
For example, a shopping agent may need to track prior purchases, compatibility requirements, and the remaining budget.
Such dependencies require agents to retain relevant history across sessions~\citep{zhang2025survey}.

Existing studies on multi-session tasks typically assume a fixed model across sessions~\citep{packer2023memgpt,wang2025agent,ouyang2026reasoningbank}.
However, in practice, orchestration and routing may lead persistent agents to swap between models to balance performance and inference cost~\citep{chen2024frugalgpt,ong2025routellm,xu2026seqroute,zhang-etal-2026-mtrouter}.
After a model swap, the \textit{recipient model} continues from the existing environment state using history generated by the \textit{source model}.
Despite the practical relevance of model swapping, its effects on performance and behavior of multi-session tasks remain underexplored.

\begin{figure}[t]
    \centering
    \includegraphics[width=\linewidth]{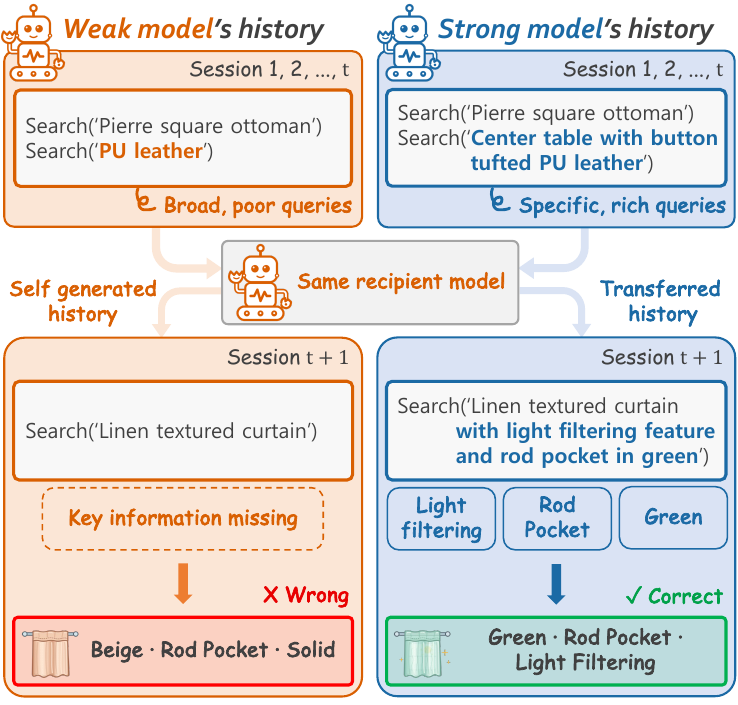}
    \caption{
    An example of behavior transfer in Web Shopping. Both pre-swap histories record the same correct purchases. With its own history, the weak model (\texttt{Qwen3.5-4B}) uses short product queries and purchases an incorrect item. With history inherited from the strong model (\texttt{Qwen3.5-27B}), the same recipient uses more detailed queries and purchases the correct item.
}
    \label{fig:behavior-case}
\end{figure}

In this work, we investigate these effects through controlled single-swap experiments across three model pairs on multi-session tasks: Web Shopping, Travel Planning, and Web Search from MemoryArena~\citep{he2026benchmarking}.
Strong-to-weak swaps generally improve recipient performance relative to the no-swap baseline, whereas weak-to-strong swaps degrade it.
This pattern persists even when the swap and no-swap histories match in success or failure at every pre-swap session, suggesting that prior-session correctness alone cannot explain these effects.

Our analysis of post-swap trajectories reveals \textit{behavior transfer}: recipients tend to adopt reasoning and action patterns from the source model's history.
These shifts occur in both strong-to-weak and weak-to-strong directions and involve query formulation, tool-use patterns, and reasoning practices such as verification.
In \cref{fig:behavior-case}, a weak recipient that typically uses simple search queries adopts more detailed queries from the stronger source model's history, succeeding on a shopping task it would fail with its own history.
These observations suggest that inherited history serves not only as a record of prior task outcomes, but also as a source of behavioral examples that shape how the recipient approaches subsequent tasks.

To examine how history editing affects behavior transfer and task performance, we compare raw-history transfer with reasoning pruning and summarization.
Both interventions reduce behavior transfer, but this does not consistently improve performance.
Mean performance across the evaluated environments is highest with raw history for strong-to-weak swaps and with summarization for weak-to-strong swaps.

To test whether this direction-dependent strategy remains effective under repeated swaps, we develop Direction- and Authorship-aware Summarization of History (DASH).
DASH passes history unchanged at strong-to-weak swaps.
At weak-to-strong swaps, it summarizes only weaker-model trajectories to retain task information while limiting their behavioral influence, leaving stronger-model trajectories intact.
On Web Shopping and Travel Planning, DASH generally outperforms raw-history transfer and direction-aware but authorship-agnostic summarization.
These results provide evidence that the history-transfer principles derived from our single-swap analysis remain effective under repeated model swaps.

Our main contributions are threefold:
\begin{itemize}
    \item We systematically study cross-model history transfer in interdependent multi-session tasks, evaluating its effects on recipient performance and behavior.
    \item We identify direction-dependent performance shifts that persist after matching earlier task outcomes, and quantify behavior transfer in recipients' reasoning and action patterns.
    \item We introduce DASH, a direction- and authorship-aware history-transfer strategy, and demonstrate its effectiveness under repeated model swaps.
\end{itemize}

\section{Model Swapping in Multi-Session Tasks}
\label{sec:formulation}

We first formalize model swapping in multi-session agents, specifying how interaction history is retained across sessions and transferred when the underlying model changes.
We then introduce raw-history transfer, swap-time reasoning pruning, and summarization as alternative ways of constructing the context exposed to the recipient model.
{Finally, we define behavior transfer and describe how we quantify it by comparing the recipient model's reasoning and action patterns under inherited and self-generated history.}

\subsection{Problem Setup}
\label{sec:multi_session_setup}

\paragraph{LLM agents in multi-session settings.}
We consider an LLM agent that completes a sequence of interdependent \emph{tasks} $x_1,\dots,x_N$ in the same environment.
Each task $x_i$ constitutes a \emph{session} $i$, during which the agent interacts with the environment over multiple \emph{steps} and produces a trajectory
\begin{equation}
    \tau_i =
    (z_{i,1},a_{i,1},o_{i,1},\dots,
    z_{i,T_i},a_{i,T_i},o_{i,T_i},y_i).
    \label{eq:session_trajectory}
\end{equation}
Here, $T_i$ is the number of interaction steps in session $i$.
At each step $t$, $z_{i,t}$, $a_{i,t}$, and $o_{i,t}$ denote the agent's reasoning, action (e.g., a tool call), and resulting observation, respectively.
The session ends with a final answer $y_i$.

{At each session, an agent uses accumulated past task inputs and trajectories to solve the current task.
Formally, interaction history $h_i$ accumulated through session $i$ is defined as:
\begin{equation}
    h_i = h_{i-1} \oplus (x_i,\tau_i)\;,
    \label{eq}
\end{equation}
where $\oplus$ denotes concatenation and $h_0=\varnothing$.
An agent model $\psi$ conditions on the preceding history $h_{i-1}$ to complete task $x_i$ and produce its session trajectory:
\begin{equation}
    \tau_i = \psi(h_{i-1}, x_i)\;.
    \label{eq}
\end{equation}
{Here, $\psi(h_{i-1},x_i)$ denotes a full session rollout rather than a single model call. At each step, the model also conditions on the reasoning, actions, and observations accumulated so far.} To manage context limits, agent memory literature often compresses interaction history or selectively retrieves it from external storage~\citep{hu2025memory}.
For simplicity of analysis, we retain history with raw trajectories across sessions by default.}

\paragraph{Model swapping and inherited context.}
{Our setting considers model swapping, where the agent model may change between sessions.}
At a model swap, a history transfer function $\Phi$ maps the retained history to the context exposed to the recipient model.
For $i\geq 1$, the context for session $i+1$ is
\begin{equation}
    c_i =
    \begin{cases}
        \Phi(h_i), & \psi_{i+1}\neq\psi_i\;, \\
        h_i, & \psi_{i+1}=\psi_i\;.
    \end{cases}
    \label{eq:context_transfer}
\end{equation}
The next session and history update then follow
\begin{equation}
    \begin{aligned}
        \tau_{i+1} &= \psi_{i+1}(c_i, x_{i+1})\;, \\
        h_{i+1} &= c_i \oplus (x_{i+1},\tau_{i+1})\;, 
    \end{aligned}
    \label{eq:swap}
\end{equation}
with $c_0=h_0=\varnothing$.
Thus, subsequent trajectories are appended to the transferred context rather than the original pre-transfer history.
The recipient continues from the environment state left by earlier sessions.
Swaps may occur repeatedly, including a return to a previously used model, as in $\psi_A\rightarrow\psi_B\rightarrow\psi_A$.\footnote{For simplicity and experimental control, we restrict model swapping within each task sequence to a fixed pair of LLMs.}

We call context generated partly or entirely from other models' interactions \emph{inherited context}.
We contrast it with \emph{self-generated context}, which would have accumulated had the recipient handled the earlier sessions itself under the same history-retention condition.
We hypothesize that inherited context affects the recipient through two factors: 
{\emph{history correctness}, since models may differ in their success or failure outcomes across pre-swap sessions,} and \emph{behavior transfer}, since context may induce the recipient to adopt the source model's reasoning patterns or tool-use habits.

\subsection{History Transfer Strategies}
\label{sec:history_transfer}

We compare three choices of $\Phi$ that differ in what they expose to the recipient.
We also distinguish these swap-time operations from a separate intervention that changes history retention after every session.

\paragraph{Raw-history transfer.}
Our baseline, $\Phi_{\mathrm{raw}}(h_i)=h_i$, passes the retained history unchanged.
It preserves task information together with the source model's reasoning and action patterns.

\paragraph{Reasoning pruning transfer.}
Let
\begin{equation}
    \tilde{\tau}_i =
    (a_{i,1},o_{i,1},\dots,a_{i,T_i},o_{i,T_i},y_i)
    \label{eq:pruned_trajectory}
\end{equation}
denote a trajectory with its reasoning traces $\{z_{i,j}\}_{j=1}^{T_i}$ removed.
The transfer function $\Phi_{\mathrm{prune}}$ replaces each retained trajectory $\tau_j$ with $\tilde{\tau}_j$ at a model swap, preserving task inputs, actions, observations, and final answers.
Pruning reduces exposure to source-model reasoning, but may also remove useful information embedded in it, such as observation summaries, intermediate results, and constraint checks.

\paragraph{Summarization-based transfer.}
Rather than removing reasoning wholesale, summarization can retain task-relevant information from it while reducing exposure to the source model's behavioral traces. 
{We summarize each unsummarized trajectory $\tau_j$ in $h_i$ independently, rather than summarizing the history as a whole. We then construct $\Phi_{\mathrm{summary}}(h_i)$ by replacing each such trajectory with its corresponding summary.}
As in \cref{eq:swap}, subsequent trajectories are appended with reasoning intact. Summarizer details are provided in \cref{app:transfer}.

\subsection{Measuring Behavior Transfer}
\label{sec:behavior_measurement}

\begingroup

History transfer determines which task information and source-model behavioral patterns the recipient observes.
We define \emph{behavior transfer} as the recipient's adoption of reasoning or action patterns from inherited history.
Such transfer may induce or suppress behaviors and affect subsequent task performance.
We quantify these behavioral shifts by comparing the recipient's behavior under inherited and self-generated context.

\paragraph{Behavioral features.}
We measure behavior using session-level features of recorded reasoning, tool calls, and final answers.
We prompt \texttt{Claude Fable 5.1}~\citep{anthropic2026fable51systemcard} to identify features from the evaluated models' trajectories and specify deterministic extraction rules.
For example, we measure query length as the mean number of words per product query in a web shopping session.

\paragraph{Measuring behavioral shifts.}
For a single swap from source $X$ to recipient $Y$, let $v_X$ and $v_Y$ denote their respective mean feature values in no-swap runs, where each model uses its own raw history.
Let $v_{X\rightarrow Y}$ denote the recipient's mean after receiving the source's history through $\Phi$.
We compute these means on aligned continuation sessions, averaging first within each task sequence and then equally across sequences.
We define the normalized behavioral shift as:
\begin{equation}
    \pi_{X\rightarrow Y}
    =
    \frac{v_{X\rightarrow Y} - v_Y}{v_X - v_Y}\;.
    \label{eq:behavior_shift}
\end{equation}
A value of zero indicates no change from the recipient baseline; one matches the source baseline.
Values between zero and one indicate a partial shift toward the source.
{Negative values indicate that the recipient's feature mean shifts from its own baseline away from the source baseline. Values above one indicate that it shifts toward the source baseline but goes past it.}
We omit normalized comparisons with insufficiently separated baselines.
Feature selection, extraction rules, session alignment, and exclusion criteria are detailed in~\cref{app:behavior}.
\par
\endgroup


\section{Understanding Cross-Model History Transfer}

\label{sec:analysis}

We study history transfer in a controlled single-swap setting, where the model changes once at a predefined session boundary.
We first analyze how model swapping affects task performance and agent behavior in~\cref{sec:analysis_performance}.
We then compare the performance and behavioral effects of raw-history transfer, reasoning pruning, and summarization in~\cref{sec:transfer_strategy}.

\begin{figure*}[t]
    \centering
    \includegraphics[width=\textwidth]{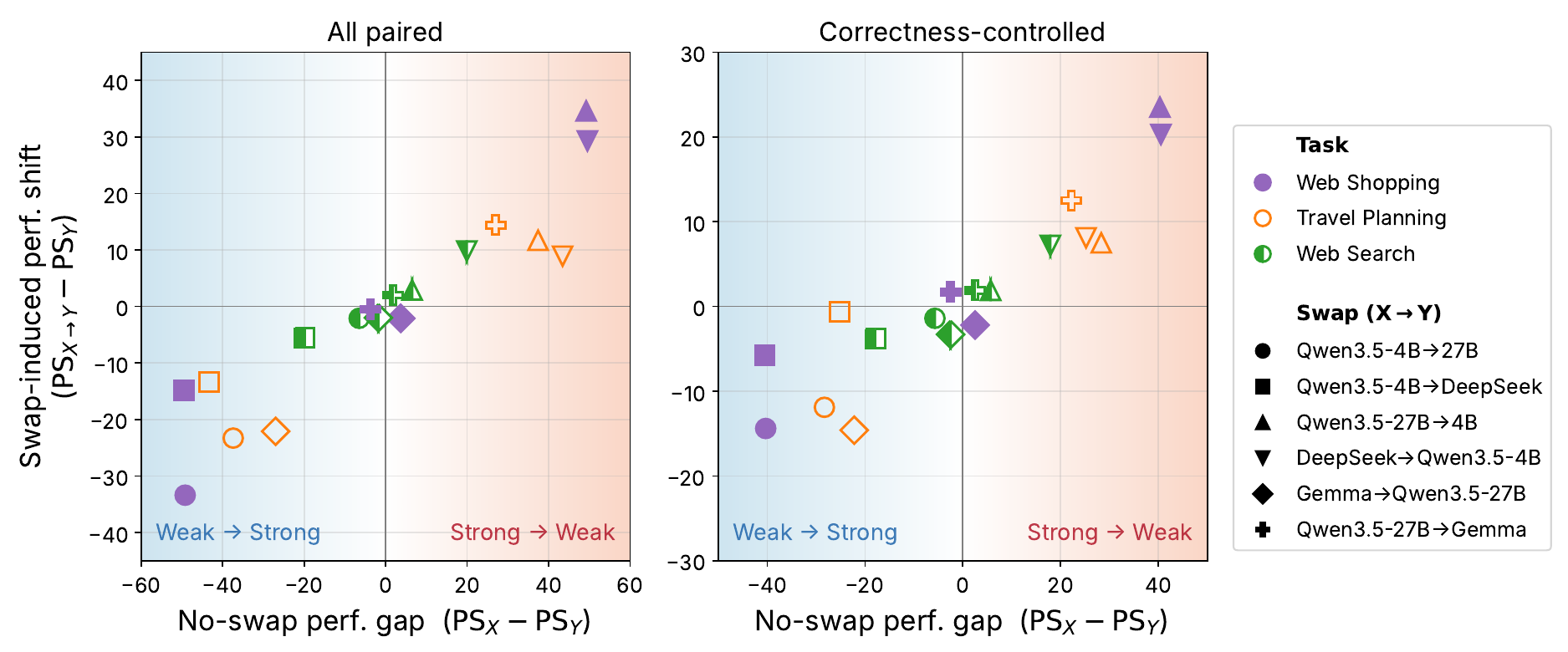}
    \caption{
        Swap-induced performance shift versus the no-swap baselines' performance gap between source $X$ and recipient $Y$, for all model pairs (left) and the correctness-controlled subset (right). Each point represents a directed swap $X \rightarrow Y$ on one agentic task, with marker shape indicating the model pair and color/fill indicating the task. The horizontal axis measures how much stronger the source is than the recipient when each model runs with its own history ($\mathrm{PS}_{X} - \mathrm{PS}_{Y}$), and the vertical axis measures how much the recipient's performance changes after inheriting the source's history, relative to running every session from scratch ($\mathrm{PS}_{X \rightarrow Y} - \mathrm{PS}_{Y}$). Blue ($x<0$) and red ($x>0$) shading indicate weak-to-strong and strong-to-weak swaps, respectively.
        For example, the purple upward-pointing triangle near $(50, 35)$ in the left panel represents a \texttt{Qwen3.5-27B} ($X$) $\rightarrow$ \texttt{Qwen3.5-4B} ($Y$) swap in Web Shopping. Its horizontal position indicates that \texttt{Qwen3.5-27B} outperforms \texttt{Qwen3.5-4B} by nearly $50$ PS when each model uses its own history, and its vertical position shows that, after inheriting \texttt{Qwen3.5-27B}'s history, \texttt{Qwen3.5-4B} gains approximately $35$ percentage points relative to the baseline in which it handles all sessions from scratch.
    }
    \label{fig:model_swapping_diff}
\end{figure*}

\subsection{Experimental Setup}
\label{sec:single_swap_setup}

\paragraph{Benchmark.}
We evaluate on three agentic environments from MemoryArena~\citep{he2026benchmarking}: Web Shopping, Travel Planning, and Web Search.
These environments are built on WebShop~\citep{yao2022webshop}, TravelPlanner~\citep{xie2024travelplanner}, and BrowseComp-Plus~\citep{chen-etal-2026-browsecomp}, respectively.
All of these are multi-session tasks: later tasks depend on earlier purchases in Web Shopping, plans in Travel Planning, and search results in Web Search.
Each instance consists of $N$ ordered subtasks $x_1,\dots,x_N$, each completed in a separate session. 

\paragraph{Swap points and comparison conditions.}
For each task sequence, we conduct separate runs with one swap after approximately 25\%, 50\%, or 75\% of its sessions.
At a swap point $k$, the source model $\psi_A$ handles the first $k$ sessions and the recipient $\psi_B$ handles the remaining $N-k$ sessions.
All comparisons, with and without a model swap and across history transfer strategies, are evaluated on the same remaining task instructions $x_{k+1},\dots,x_N$.

\paragraph{Metrics.}
We report MemoryArena's \emph{Task Progress Score} (PS)~\citep{he2026benchmarking}.
Each subtask receives a binary score according to the environment-specific success criteria.
For each task sequence with a swap after session $k$, post-swap PS is the fraction of the remaining $N-k$ subtasks completed successfully. For a swap from $X$ to $Y$, we denote the recipient's post-swap score as $\mathrm{PS}_{X\rightarrow Y}$, and its no-swap baseline score as $\mathrm{PS}_Y$.
Within each environment, we average scores across task sequences, giving each sequence equal weight regardless of length.

\paragraph{LLM pairs.}
We use \texttt{Qwen3.5-4B}, \texttt{Qwen3.5-27B}~\citep{qwen35blog}, \texttt{Gemma-4-31B-it}~\citep{team2026gemma}, and \texttt{DeepSeek-V4.1-Flash}~\citep{xu2026deepseekv4-1}.
We abbreviate the latter two models as \texttt{Gemma} and \texttt{DeepSeek}.
We evaluate three pairs: \{\texttt{Qwen3.5-4B}, \texttt{Qwen3.5-27B}\}, \{\texttt{Qwen3.5-4B}, \texttt{DeepSeek}\}, and \{\texttt{Qwen3.5-27B}, \texttt{Gemma}\}.

\paragraph{Agent configuration.}
Within each history-transfer condition, we keep the agent harness fixed across swaps, changing only the underlying model.
In Web Search, inputs exceeding 128k tokens are compacted by summarizing earlier content, independently of the swap-time history-transfer methods.
Further implementation details are provided in \cref{app:exp_setting}.



\subsection{Model Swapping Analysis}
\label{sec:analysis_performance}

\paragraph{Performance shifts with and without correctness control.}
\cref{fig:model_swapping_diff} plots the recipient's performance shift, $\mathrm{PS}_{X \rightarrow Y}-\mathrm{PS}_{Y}$, against the source–recipient performance gap without swapping, $\mathrm{PS}_{X}-\mathrm{PS}_{Y}$, across model pairs and agentic tasks.
The left panel shows results over all paired runs. 
The right panel controls for prior-session correctness, retaining only paired no-swap and swap runs with the same correct or incorrect outcome in \emph{every} pre-swap session.

The two quantities are strongly correlated in both settings, with Pearson correlations of $r=0.93$ without correctness control and $r=0.91$ with correctness control.
In both settings, strong-to-weak swaps generally improve performance relative to the recipient's self baseline, whereas weak-to-strong swaps generally degrade it.
The shifts are smaller under correctness control, but the directional pattern persists, indicating that prior-session correctness alone does not fully account for the observed effects.
Matching correctness does not make the histories or environment states identical; we next examine behavioral shifts as additional evidence.

\paragraph{Behavior transfer.}
\label{sec:analysis_behavior}
\begin{figure}[t]
    \centering
    \includegraphics[width=\linewidth]{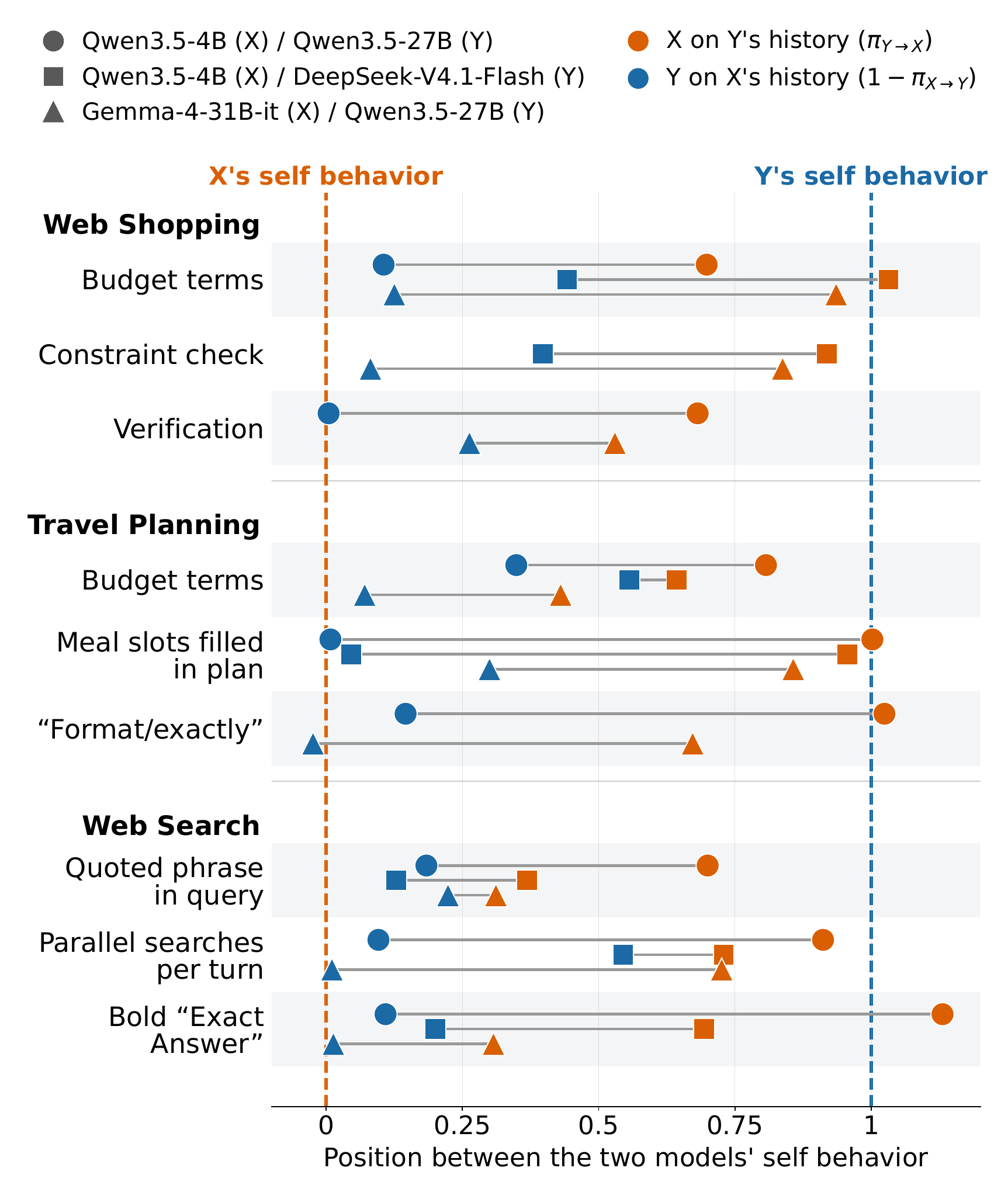}
    \caption{Behavior transfer after model swapping.
    Recipients tend to adopt behavioral patterns from the source model’s history.
    For each behavioral feature and model pair $(X,Y)$, position $0$ and $1$ on the horizontal axis represent the baseline behavior feature values of $X$ and $Y$, respectively, without model swap.
    Orange markers show $X$ with $Y$’s history, and blue markers show $Y$ with $X$’s history.
    Thus, orange markers near $1$ and blue markers near $0$ show that the behaviors of the recipient are similar to those of the source model baseline.
    The plotted positions are $\pi_{Y\rightarrow X}$ for orange markers and $1-\pi_{X\rightarrow Y}$ for blue markers (\cref{eq:behavior_shift}).
    Points are omitted when the two baseline models have nearly identical feature values.
    \cref{fig:websearch-behavior-compaction} compares behavior before and after compaction, and \cref{tab:behavior-full} provides extended results.}
    \label{fig:behavior-transfer-main}
\end{figure}
\cref{fig:behavior-transfer-main} shows that the recipient's reasoning behavior shifts toward the source's in Web Shopping and Travel Planning.
With its own history, \texttt{Qwen3.5-27B} mentions budget terms and verification more often than \texttt{Qwen3.5-4B} in Web Shopping, and mentions output-format constraints and earlier sessions more often in Travel Planning.
These feature values decrease when \texttt{Qwen3.5-27B} inherits \texttt{Qwen3.5-4B}'s history, nearly reaching the smaller model's levels for budget terms and verification.
Conversely, \texttt{Qwen3.5-4B} exhibits more of these behaviors after inheriting \texttt{Qwen3.5-27B}'s history.
Inherited context can therefore induce as well as suppress behaviors.
These shifts are consistent with behavior transfer, although they do not by themselves establish its causal contribution to the performance changes.

Similar shifts occur between \texttt{Qwen3.5-27B} and \texttt{Gemma}, showing that behavior transfer is not limited to models with substantially different capabilities.
Selected features show weaker shifts in Web Search, where document-heavy histories and context compaction may limit exposure to source-model reasoning patterns.

\begin{figure}[t]
    \centering
    \includegraphics[width=\linewidth]{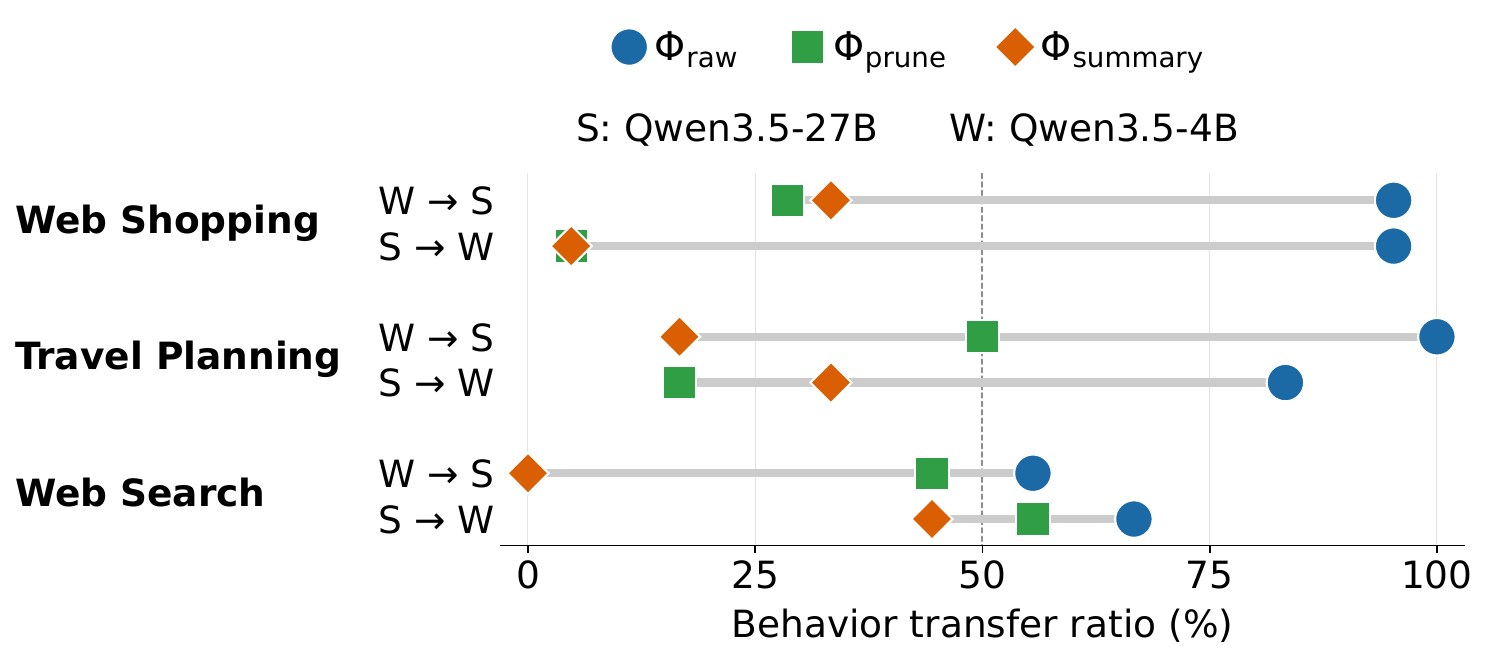}
    \caption{
    Behavior transfer is strongest under raw-history transfer, substantially reduced by reasoning pruning, and tends to be reduced further by summarization.
    {The horizontal axis shows the percentage of behavioral features for which the recipient is closer to the source model's no-swap baseline than to its own ($\pi_{X\rightarrow Y}>0.5$), with both baselines measured under self-generated raw history.}
    For each environment and swap direction, percentages are computed over features with valid normalized behavioral shifts.
    Feature-level results are provided in \cref{tab:behavior-full,tab:behavior-prune,tab:behavior-summary}, and \cref{fig:median_4b_27b_extended} shows results for additional model pairs.
    }
    \label{fig:median_4b_27b}
\end{figure}

\subsection{Comparison of History Transfer Strategies}
\label{sec:transfer_strategy}
\paragraph{Behavior transfer under pruning and summarization.}
{We next compare $\Phi_{\mathrm{prune}}$ and $\Phi_{\mathrm{summary}}$ with $\Phi_{\mathrm{raw}}$ for the same pre-swap histories. All swap-time summaries are generated by \texttt{Qwen3.5-27B}. For the \texttt{Qwen} pair, \cref{fig:median_4b_27b} plots the \emph{behavior transfer ratio} for each environment and swap direction. For each valid behavioral feature, we compare the recipient's post-swap mean with the source and recipient no-swap baselines. The behavior transfer ratio is the percentage of these features whose post-swap means are closer to the source baseline than to the recipient baseline ($\pi_{X\rightarrow Y}>0.5$). Higher values therefore indicate a larger share of source-like behavioral features. Pruning and summarization both lower this ratio relative to raw-history transfer across all three environments and both swap directions. These results indicate that editing the transferred history can limit the recipient's adoption of source-model reasoning and action patterns.}

\begin{figure}[t]
    \centering
    \includegraphics[width=\linewidth]{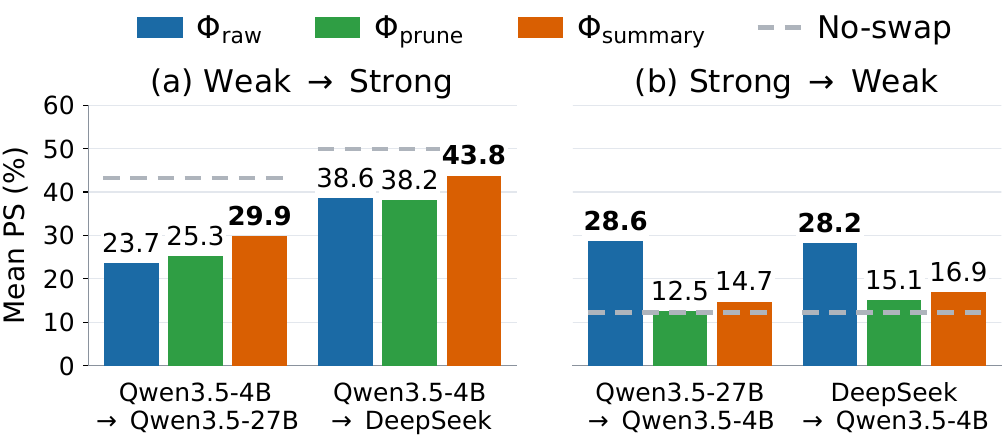}
    \caption{Mean post-swap performance ($\mathrm{PS}_{X\rightarrow Y}$) across environments with different history transfer strategies.
    An effective history transfer strategy depends on the model swap direction: summarization achieves the highest mean PS for weak-to-strong swaps, whereas raw-history transfer performs best for strong-to-weak swaps.
    Scores are averaged equally over Web Shopping, Travel Planning, and Web Search for (a) weak-to-strong and (b) strong-to-weak swaps.
    \emph{No-swap} is the recipient model's baseline performance without a model swap ($\mathrm{PS}_{Y}$).
    Bold values indicate the highest mean PS within each group.
    Full results are reported in \cref{tab:memoryarena-model-swapping}.}
    \label{fig:swap_intervention}
\end{figure}

\paragraph{Task performance under pruning and summarization.}
Despite their shared effect on behavior transfer, pruning and summarization differ in their performance implications. \cref{fig:swap_intervention} reports the unweighted mean post-swap PS across the three environments, with environment-level results in \cref{tab:memoryarena-model-swapping}. For weak-to-strong swaps, pruning provides limited and inconsistent improvements. As shown in \cref{fig:swap_intervention}(a), it modestly raises mean PS for \texttt{Qwen3.5-4B}$\to$\texttt{Qwen3.5-27B}, from 23.7 to 25.3, but slightly lowers it for \texttt{Qwen3.5-4B}$\to$\texttt{DeepSeek}, from 38.6 to 38.2. Summarization outperforms both raw-history transfer and pruning for both pairs, reaching 29.9 and 43.8, respectively, although performance remains below the recipients’ no-swap baselines.

For strong-to-weak swaps, raw-history transfer achieves the highest mean PS for both pairs. In \cref{fig:swap_intervention}(b), pruning and summarization lower mean PS from 28.6 to 12.5 and 14.7 for \texttt{Qwen3.5-27B}$\to$\texttt{Qwen3.5-4B}, and from 28.2 to 15.1 and 16.9 for \texttt{DeepSeek}$\to$\texttt{Qwen3.5-4B}. Thus, attenuating behavior transfer does not consistently improve performance. These losses may reflect the removal of useful task information, beneficial behavioral guidance, or both. Reasoning traces can contain observation summaries, intermediate results, and constraint checks, alongside examples of how the source model approaches the task. Summarization may retain some of this information more effectively than pruning, but neither intervention fully preserves the benefits of the stronger model’s trajectories in these comparisons.

\paragraph{Implications for transfer strategy.}
These comparisons motivate preserving useful behavioral guidance rather than suppressing behavior transfer indiscriminately. In the reported three agentic environments, they favor retaining raw history for strong-to-weak swaps and using summarization for weak-to-strong swaps.
In \cref{sec:exp}, we extend this direction-dependent approach to histories containing trajectories from both models.


\section{Evaluation under Repeated Swapping}
\label{sec:exp}

We extend the single-swap transfer strategy in  \cref{sec:analysis} to repeated swap settings, where models may change repeatedly within a task sequence.

\subsection{Experimental Settings}
We evaluate history transfer strategies in Web Shopping and Travel Planning. We exclude Web Search because frequent history compaction can obscure the effects of swap-time history transfer. We describe the remaining experimental settings below.

\paragraph{LLM pairs.}
For the routing experiments, we focus on swaps between models with different capabilities.
We evaluate four pairs\footnote{Major closed-source reasoning models, including current Claude, GPT, and Gemini, do not expose raw reasoning traces through their standard APIs. This limits our experiments using raw reasoning traces.}: \{\texttt{Qwen3.5-4B}, \texttt{Qwen3.5-27B}\}, \{\texttt{Qwen3.5-4B}, \texttt{DeepSeek-V4.1-Flash}\}, \{\texttt{Gemma-4-E4B-it}, \texttt{Gemma-4-31B-it}\}, and \{\texttt{Gemma-4-E4B-it}, \texttt{DeepSeek-V4.1-Flash}\}.
Within each pair, we denote the first model as the weaker model $\psi_W$ and the second as the stronger model $\psi_S$.

\paragraph{Repeated swaps.}
As an extension of the single-swap setting in \cref{sec:analysis}, we also consider task sequences in which control alternates between the two models several times. Within a single sequence of $N$ sessions, we swap the acting model at approximately 25\%, 50\%, and 75\% of the sessions, so the sequence consists of four alternating blocks. We test both orders: \emph{strong-first} ($\psi_S \to\psi_W \to \psi_S \to \psi_W$) and \emph{weak-first} ($\psi_W \to \psi_S \to \psi_W \to \psi_S$). 

\paragraph{History transfer strategies.}
{We introduce \emph{Direction- and Authorship-aware Summarization of History} (DASH) for repeated model swaps. At weak-to-strong swaps, DASH summarizes only previously unsummarized trajectories authored by the weaker model, while preserving the stronger model's trajectories verbatim. Existing summaries are retained, and the history is passed
unchanged at strong-to-weak swaps.}

{As an authorship-agnostic baseline, we consider \emph{Directional Summarization of History} (DiSH), which summarizes all remaining raw trajectories at weak-to-strong swaps, regardless of their authorship.}

\paragraph{Metrics.}
We report PS as introduced in \cref{sec:analysis}, computed over all sessions following the first model swap. We exclude the initial block because all three history transfer strategies share the same history before the first swap.

\begin{table}[t]
    \centering
    \small
    \setlength{\tabcolsep}{3pt}
    \renewcommand{\arraystretch}{1.08}
    \resizebox{\linewidth}{!}{%
    \begin{tabular}{@{}llrrrr@{}}
        \toprule
        &
        \shortstack{\textbf{History}\\\textbf{transfer}}
        & \shortstack{
            \texttt{Qwen3.5-4B}\\
            \texttt{Qwen3.5-27B}
        }
        & \shortstack{
            \texttt{Qwen3.5-4B}\\
            \texttt{DeepSeek}
        }
        & \shortstack{
            \texttt{Gemma-4-E4B}\\
            \texttt{Gemma-4-31B}
        }
        & \shortstack{
            \texttt{Gemma-4-E4B}\\
            \texttt{DeepSeek}
        } \\
        \midrule

        \multicolumn{6}{@{}l}{\textit{Weak-first} ($\psi_W \to \psi_S \to \psi_W \to \psi_S$)} \\ \midrule
        \multirow{3}{*}{\rotatebox[origin=c]{90}{Web}}
        & $\Phi_{\mathrm{raw}}$  & 23.5 & 32.5 & 24.8 & 27.5 \\
        & $\Phi_{\mathrm{DiSH}}$ & 27.2 & 39.5 & 30.5 & 35.7 \\
        & $\Phi_{\mathrm{DASH}}$ & 32.3 & 39.8 & 31.7 & 35.5 \\
        \cmidrule{2-6}
        \multirow{3}{*}{\rotatebox[origin=c]{90}{Travel}}
        & $\Phi_{\mathrm{raw}}$  & 29.3 & 32.2 & 18.6 & 27.5 \\ 
        & $\Phi_{\mathrm{DiSH}}$ & 31.3 & 47.8 & 24.8 & 40.9 \\
        & $\Phi_{\mathrm{DASH}}$ & 34.3 & 48.8 & 24.8 & 46.1 \\

        \midrule
        \multicolumn{6}{@{}l}{\textit{Strong-first} ($\psi_S \to \psi_W \to \psi_S \to \psi_W$)} \\ \midrule
        \multirow{3}{*}{\rotatebox[origin=c]{90}{Web}}
        & $\Phi_{\mathrm{raw}}$  & 45.6 & 44.8 & 32.0 & 33.1 \\ 
        & $\Phi_{\mathrm{DiSH}}$ & 39.5 & 42.0 & 33.6 & 34.5 \\
        & $\Phi_{\mathrm{DASH}}$ & 50.3 & 48.5 & 34.0 & 33.7 \\
        \cmidrule{2-6}
        \multirow{3}{*}{\rotatebox[origin=c]{90}{Travel}}
        & $\Phi_{\mathrm{raw}}$  & 33.1 & 37.2 & 18.1 & 29.8 \\ 
        & $\Phi_{\mathrm{DiSH}}$ & 30.7 & 42.6 & 18.8 & 34.1 \\
        & $\Phi_{\mathrm{DASH}}$ & 34.1 & 42.5 & 22.7 & 33.4 \\
        \bottomrule
    \end{tabular}
    }
    \caption{
        Performance of history transfer strategies under repeated model swaps on Web Shopping (Web) and Travel Planning (Travel). The proposed DASH transfer strategy achieves the highest PS in most evaluated settings.
        Values report PS (\%) over all sessions following the first swap, for weak-first and strong-first model orders. Each model-pair column lists the weaker model $\psi_W$ above the stronger model $\psi_S$. 
        \texttt{DeepSeek} denotes \texttt{DeepSeek-V4.1-Flash}, and all \texttt{Gemma} models are the instruction-tuned (\texttt{-it}) variants.
    }
    \label{tab:multiswap}
\end{table}

\subsection{Results}


\paragraph{Performance under repeated swaps.}
\cref{tab:multiswap} shows that direction-aware summarization generally improves PS over raw-history transfer, supporting its effectiveness under repeated model swaps. DASH achieves higher PS than raw transfer in general and outperforms DiSH in 11 settings, typically by a modest margin. These gains suggest that preserving stronger-model trajectories, rather than summarizing all remaining raw trajectories at weak-to-strong routing, can help retain useful task information and behavioral guidance for subsequent sessions.

\begin{figure}[t]
    \centering
    \includegraphics[width=0.95\linewidth]{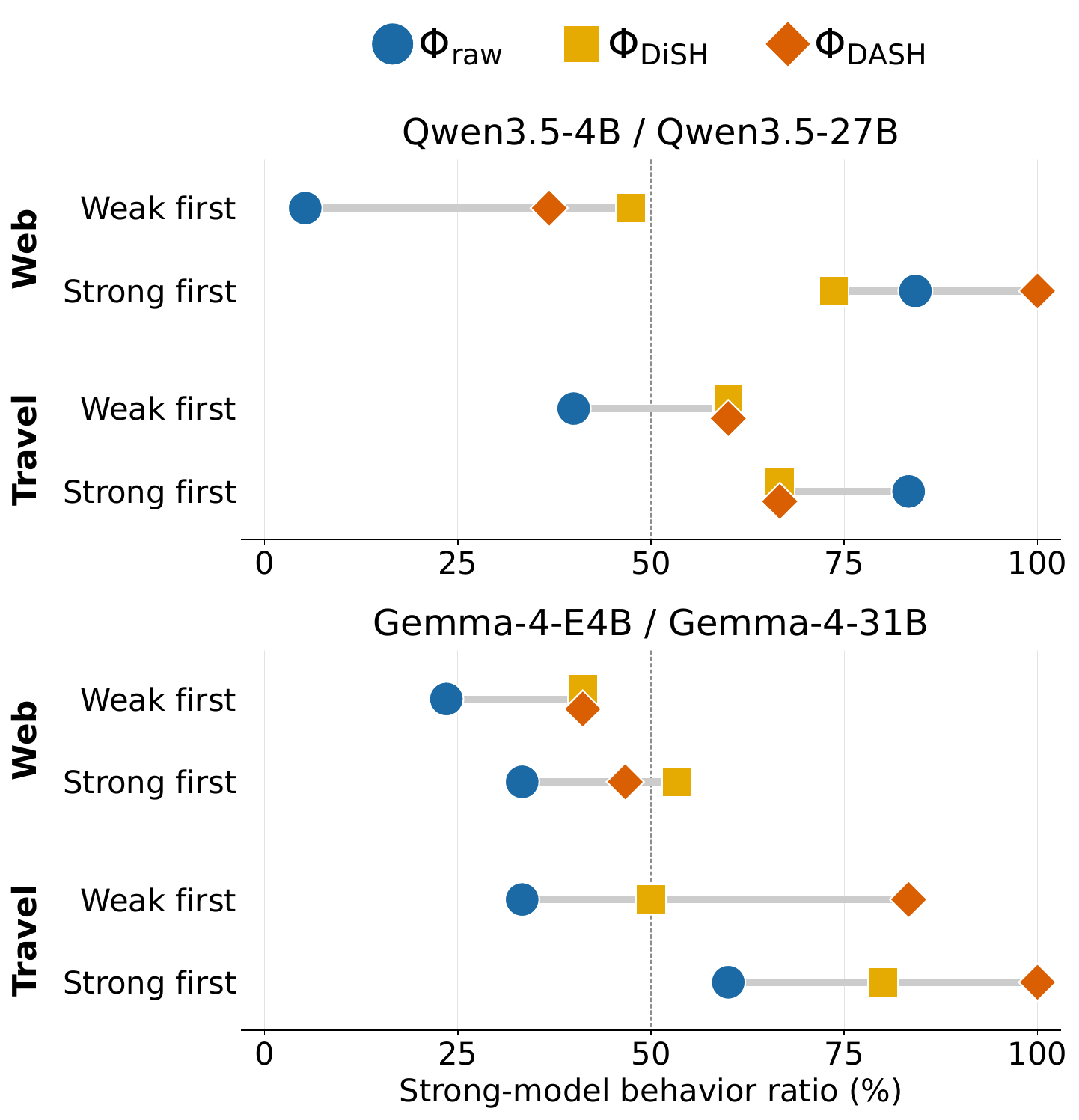}
    \caption{
    Direction-aware summarization shifts post-swap behavior toward the strong model, especially under weak-first routing. The horizontal axis shows the percentage of behavioral features whose post-swap means are closer to the strong model's no-swap baseline than to the weak model's ($(v_{\mathrm{post}}-v_W)/(v_S-v_W)>0.5$). Feature means pool sessions after the first swap regardless of the acting model, and percentages are computed over features with valid normalized shifts. \cref{fig:multiple_swap_full} shows results for additional model pairs.
    }
    \label{fig:multiple_swap_qwen}
\end{figure}
\paragraph{Behavior under repeated swaps.}
For each behavioral feature, we compute its post-swap mean $v_{\mathrm{post}}$ by pooling sessions from both models after the first swap. We compare this mean with $v_W$ and $v_S$, the corresponding weak- and strong-model no-swap baselines.
\cref{fig:multiple_swap_qwen} reports the \emph{strong-model behavior ratio}, the percentage of valid behavioral features for which post-swap behavior is closer to that of the strong model, i.e., $(v_{\mathrm{post}}-v_W)/(v_S-v_W)>0.5$.
DiSH and DASH generally increase this ratio under weak-first routing. With raw history, the ratio is already higher under strong-first than weak-first routing, consistent with the influence of earlier strong-model trajectories. DASH's gains over DiSH do not consistently coincide with a higher ratio. We hypothesize that these gains partly reflect task information preserved in stronger-model trajectories, whose usefulness may vary across tasks and recipient models.

\paragraph{Choice of the starting model.}
Strong-first routing generally yields higher PS with raw history and remains preferable in most Web Shopping settings under DiSH and DASH.
In Travel Planning, however, weak-first yields higher PS for every evaluated pair under both summarization strategies.
The preferred routing order thus depends on the environment and transfer strategy.
These comparisons reflect model assignments throughout the sequence as well as the initial history.

\section{Related Work}
Our work connects research on memory for multi-session agents, model routing and handoffs, and contextual influences on LLM behavior.
Memory systems retain and reuse experience~\citep{hu2025memory,ouyang2026reasoningbank}, while routing selects models to balance performance and cost~\citep{chen2024frugalgpt,ong2025routellm,feng2025graphrouter}.
Recent handoff studies examine switches within ongoing dialogues or tasks~\citep{khraishi2026evaluating,ganz2026handoff}.
Context studies show that prior examples and trajectories shape subsequent behavior~\citep{brown2020language,menon2026inherited}.
We study swaps across interdependent sessions, where inherited history carries both task information and examples of source-model behavior.
We quantify the transfer of reasoning and action patterns and develop history-transfer strategies that account for swap direction and trajectory authorship.
See \cref{app:related_work} for a detailed discussion.

\section{Conclusion}

We studied model swapping in multi-session LLM agents and observed behavior transfer from inherited history. Under raw-history transfer, strong-to-weak swaps generally improve performance and weak-to-strong swaps degrade it relative to self-generated history, even after matching prior-session correctness. Our comparisons support direction-dependent guidelines for history transfer: retaining raw history for strong-to-weak swaps and using summarization for weak-to-strong swaps. We extend these guidelines to repeated model swaps through DASH, which summarizes weaker-model trajectories at weak-to-strong swaps while preserving stronger-model trajectories. DASH achieves higher PS than raw-history transfer in all evaluated settings and outperforms DiSH in most settings, typically by a modest margin. These findings support accounting for both swap direction and trajectory authorship when transferring history in multi-session LLM agents.

\section*{Limitations}


{
We evaluate fixed model pairs with swaps at predefined session boundaries to enable controlled comparisons of history-transfer strategies. Our repeated-swap experiments focus on Web Shopping and Travel Planning and consider both strong-first and weak-first orders. This setup examines history transfer under a given model sequence, rather than jointly optimizing model selection and swap timing. Extending DASH to larger model pools and adaptive swap schedules remains future work.}

{Behavioral features are proposed with LLM assistance and then measured using fixed, deterministic extraction rules applied consistently across models and conditions. These rules support reproducible comparisons, but the resulting measurements capture a selected set of observable reasoning, tool-use, and output patterns rather than all aspects of agent behavior.}

{Throughout the experiments, agents receive retained interaction history directly in context, without retrieving past interactions from an external memory system. The evaluation therefore focuses on transforming the history available at a swap rather than selecting which past interactions to retrieve. The effectiveness of DASH when combined with retrieval-based memory remains to be evaluated, particularly how authorship-aware summarization interacts with the selection of past interactions for subsequent tasks.}

{Finally, reasoning-level behavioral analysis and pruning require access to raw reasoning traces. Consequently, closed-source models whose APIs do not expose these traces were not evaluated as agents in these experiments.
}

\bibliography{custom}

\clearpage
\appendix
\section*{Appendix}
\crefalias{section}{appendix}
\crefalias{subsection}{appendix}

\numberwithin{figure}{section}
\numberwithin{table}{section}

\crefname{figure}{Figure}{Figures}
\crefname{table}{Table}{Tables}

\section{In-Depth Analysis of Related Work}
\label{app:related_work}

\paragraph{Multi-session agentic task and memory.}
Multi-session agentic tasks require agents to act on an environment across interdependent sessions, motivating benchmarks that evaluate continuity and sustained task execution~\citep{zheng2025lifelongagentbench,he2026benchmarking, li2026agencybench}.
Solving these tasks requires retaining and reusing relevant history across interactions, commonly supported by agent memory approaches~\citep{hu2025memory}.
Existing approaches organize information for storage and retrieval~\citep{xu2025amem}, distill prior trajectories into reusable procedures and reasoning strategies~\citep{fang2026memp,ouyang2026reasoningbank}, or adapt the memory architecture itself~\citep{zhang2026memevolve}.
These approaches primarily focus on constructing, updating, and reusing memory across successive tasks.
In contrast, we study settings where the underlying model changes during a multi-session task.
Rather than designing memory updates for every session, we focus on the swap itself: how existing history affects the recipient model's performance and behavior, and how to transfer that history effectively.


\paragraph{Model routing and model swaps.}
{LLM routing provides a situation where agents may switch models, selecting from a model pool to balance task performance and inference cost~\citep{chen2024frugalgpt,ong2025routellm}.
Routing across dialogue turns and agent steps involves repeated model selection~\citep{xu2026seqroute,zhang-etal-2026-mtrouter}, with a primary focus on which model to use at each decision point rather than how inherited history affects the recipient model or should be transferred.
Most closely related to our setting,~\citet{khraishi2026evaluating} study model switching within multi-turn dialogue and show directional performance drift.
\citet{ganz2026handoff} study model handoffs within coding-agent trajectories and show that the preferred history-transfer strategy depends on swap direction.
These works study continuation within an ongoing dialogue or task. In contrast, we study model swaps across interdependent sessions, where prior history is itself required for subsequent tasks.
We directly analyze the transfer of source-model action patterns and develop direction-dependent history-transfer guidelines, including factual summarization to preserve task information while limiting unwanted behavior transfer.}

\paragraph{Effect of context on LLM behavior.}
In-context examples can guide LLM behavior without parameter updates~\citep{brown2020language}, and their format, ordering, and relevance affect its performance~\citep{min2022rethinking,lu2022fantastically,shi2023large}.
Dialogue history also shapes subsequent behavior: models remain anchored to early assumptions and struggle to incorporate corrective feedback~\citep{laban2026llms,choi2025place}. {Agents may imitate their own earlier responses or retrieved experiences~\citep{wan2026mitigating,xiong2026memory}. They can also inherit goal deviations from trajectories generated by other models~\citep{menon2026inherited}.}
We examine these contextual influences across model swaps in multi-session tasks, where inherited history provides both information needed for task continuity and examples of source-model behavior.
Beyond measuring performance changes, we quantify whether recipients adopt the source model's reasoning and action patterns.
We also compare histories with matching success or failure outcomes in every pre-swap session, showing that prior-session correctness alone does not fully explain the observed performance changes.

\clearpage

\section{Detailed Experimental Settings}
\label{app:exp_setting}

\subsection{Benchmark and Evaluation}
\label{app:benchmark_detail}
We use all task sequences from the five MemoryArena environments~\citep{he2026benchmarking}, summarized in \cref{tab:data}. Each sequence contains $N$ interdependent subtasks, each completed in a separate session. The agent receives the retained interaction history directly in its context, as described in \cref{sec:formulation}, without an external memory system or retrieval of past interactions. We use the environment tools and interaction protocols of the MemoryArena harness, keeping them fixed across models and conditions.

\begin{table}[h]
    \centering
    \small
    \setlength{\tabcolsep}{4pt}
    \begin{tabular}{@{}lrrrr@{}}
        \toprule
        Environment & Tasks & Sessions & $N$ & Swap points \\
        \midrule
        Web Shopping    & 150 &   900 & 6       & 450 \\
        Travel Planning & 270 & 1{,}869 & 5--8  & 773 \\
        Web Search      & 221 & 1{,}641 & 4--16 & 633 \\
        Math            &  40 &   354 & 2--16   & 115 \\
        Physics         &  20 &    86 & 2--12   &  47 \\
        \bottomrule
    \end{tabular}
    \caption{MemoryArena statistics used in our experiments. Tasks and Sessions count task sequences and individual sessions, respectively. $N$ is the number of sessions per sequence. The last column reports the number of evaluated swap positions.}
    \label{tab:data}
\end{table}

\paragraph{Web Shopping~\citep{yao2022webshop}.}
Each task requires purchasing six mutually compatible products, one per session, under a shared budget. The agent issues one WebShop action per turn, either \texttt{search[$\cdot$]} or \texttt{click[$\cdot$]}, and ends the session by clicking \texttt{Buy Now}. Each purchase must be compatible with earlier purchases. Following the MemoryArena evaluator, a session succeeds when the purchased product exactly matches the target ASIN.

\paragraph{Travel Planning~\citep{xie2024travelplanner}.}
Each task begins with a finalized itinerary for a base traveler and adds one traveler per session. New requirements refer to earlier travelers' plans through shared choices, such as joining another traveler's dinner, or relative constraints, such as a restaurant priced within 30\% of another traveler's breakfast. The agent searches the TravelPlanner database for flights, restaurants, accommodations, attractions, and distances, then produces a day-by-day plan in a fixed format. A session succeeds when all six slots, such as breakfast, lunch, dinner, accommodation, transportation, and attraction, match the gold plan with string similarity of at least 0.7 on every day of the gold plan, following the MemoryArena evaluator.

\paragraph{Web Search~\citep{chen-etal-2026-browsecomp}.}
Each task decomposes a BrowseComp-Plus question into subqueries that successively add constraints, with the original question posed in the final session. A retrieval tool searches the BrowseComp-Plus corpus and returns the top 10 documents with snippets truncated to 512 tokens. We grade each session with \texttt{gpt-5-mini} using the BrowseComp grader prompt provided by MemoryArena, which compares the full response with the gold answer. Context compaction is described in \cref{app:inference}.

\paragraph{Formal Reasoning.}
Math and Physics decompose a research result into ordered lemmas and propositions, with later sessions building on earlier results. Each session provides the background and current problem for a natural-language answer. The model can answer directly or call \texttt{reasoning}, which forwards its argument to the same model in a fresh conversation without session history, using a fixed solving prompt.\footnote{The tool uses the instruction ``\textit{Solve the following math problem carefully. Provide concise reasoning, check your work, and state the final answer clearly.}''} The tool provides no external information, and its output is appended to the session context. A direct answer ends the session. 

Each session has at most three turns, and if the third turn invokes the tool, its output becomes the final answer. We grade answers with \texttt{gpt-5-mini} using MemoryArena's equivalence prompt. We also provide the problem statement, which the original grader omits, because equivalence can depend on the question being answered.
In Math and Physics, the derivation can be part of the answer itself, so reasoning cannot be cleanly separated from answer content. We therefore do not evaluate reasoning-pruned history or swap-time reasoning pruning in these environments. We provide the results of formal reasoning in \cref{app:reasoning_results}. 

\subsection{LLM Inference Settings}
\label{app:inference}

We serve \texttt{Qwen3.5-4B}, \texttt{Qwen3.5-27B}~\citep{qwen35blog}, and \texttt{Gemma-4-31B-it}~\citep{team2026gemma} locally with vLLM~\citep{kwon2023efficient}, using each model's tool-call and reasoning parsers. We access \texttt{DeepSeek-V4.1-Flash}~\citep{xu2026deepseekv4-1} through OpenRouter with DeepInfra as the provider. \cref{tab:inference} summarizes the generation settings, following the MemoryArena configuration for each environment. We report the results with a single run.

\begin{table}[h]
    \centering
    \small
    \setlength{\tabcolsep}{3pt}
    \begin{tabular}{@{}lrrr@{}}
        \toprule
        Environment & Max tokens & Reasoning budget & Max steps \\
        \midrule
        Web Shopping    & 4{,}096  & 3{,}584  & 20 \\
        Travel Planning & 32{,}768 & 22{,}938 & 30 \\
        Web Search      & 15{,}000$^\dagger$ & 1{,}024 & 35 \\
        Math, Physics   & 8{,}192  &  --     & 3 \\
        \bottomrule
    \end{tabular}
    \caption{Generation settings. Max tokens is the limit per model step, except for Web Search, where it is a session-level budget ($^\dagger$). Reasoning budgets apply per step, with reasoning mode disabled for Math and Physics.}
    \label{tab:inference}
\end{table}

\paragraph{Output limits.}
For Math and Physics, the 8{,}192-token limit applies to every model call, including calls inside the \texttt{reasoning} tool. Truncated answers are graded as is, and a session is marked \emph{cut} if any call reaches this limit. Empty answers are graded as incorrect.

\paragraph{Decoding.}
We use greedy decoding except in Web Search, where it leads models to repeat the same query in consecutive tool calls. For Web Search, we use temperature 0.6, top-$p$ 0.95, and top-$k$ 20.

\paragraph{Reasoning retention.}
When reasoning mode is enabled, the model's separate reasoning traces remain in its context within a session. Under raw-history transfer, these traces are also retained across sessions. Under reasoning-pruned history, they are removed at each session boundary. The API provider for \texttt{DeepSeek-V4.1-Flash} drops reasoning from earlier answer turns in the input, so we insert it into the assistant message between \texttt{[Thinking]} and \texttt{[/Thinking]} tags. This preserves the reasoning content available to the locally served models.

\paragraph{Context management.}
We cap retained history at 120k tokens, including reasoning, by removing the oldest sessions in full. Within a session, if the context would exceed the model's window, we remove the oldest reasoning first and then the oldest turns, always preserving the system prompt and current task instruction.

Web Search instead compacts the input to a model call when it exceeds 128k tokens. Completed sessions are summarized first, followed by older turns of the current session, until the input is reduced to approximately 80\% of the trigger length. The earlier content is replaced by a summary of at most 4{,}096 tokens, using the prompt in \cref{fig:prompt-compaction}. The acting model writes the summary with reasoning disabled, except that \texttt{Qwen3.5-27B} summarizes for \texttt{DeepSeek-V4.1-Flash} to limit API costs. If summarization fails, we remove the oldest content until the input is at most 120k tokens. This compaction affects only the current call's input. Later sessions inherit the uncompacted trajectory, subject to the same 120k-token retained-history limit.

\subsection{Summarization-based History Transfer Strategies}
\label{app:transfer}

We implement $\Phi_{\mathrm{summary}}$ by summarizing each unsummarized trajectory in the retained history independently. Original requests and existing summaries are retained verbatim. The summarizer receives the system prompt, the selected session's request and trajectory as exposed to the acting model, and the summarization instruction in \cref{fig:prompt-summary} as the final user turn. All summaries are written by \texttt{Qwen3.5-27B}. In the single-swap setting, the recipient receives a single hand-off user message containing each retained session's original request followed by its summary. Subsequent trajectories are appended with reasoning intact.

For multiple swaps, $\Phi_{\mathrm{DiSH}}$ applies this operation to all unsummarized sessions at weak-to-strong swaps. $\Phi_{\mathrm{DASH}}$ additionally considers session authorship and summarizes only unsummarized sessions from $\psi_W$, retaining trajectories from $\psi_S$. Both strategies preserve existing summaries and pass the retained history unchanged at strong-to-weak swaps.

\subsection{Behavior Transfer Analysis}

We describe the feature definitions and evaluation protocol for behavior transfer analysis. We provide extended tabulated results in \cref{tab:behavior-full}.

\paragraph{Feature selection.}
We use \texttt{Claude Fable 5.1} to help propose behavioral features by inspecting trajectories from all models in Web Shopping, Travel Planning, and Web Search. Each feature is then measured using a fixed, deterministic rule over recorded reasoning, tool calls, or final answers, with the same rule applied across models and conditions.
We split each environment's task sequences into discovery and held-out halves. Feature screening uses the discovery half. We retain features that are defined in at least 80\% of task sequences, have split-half reliability of at least 0.5 across sessions, and distinguish the no-swap history behavior of \texttt{Qwen3.5-4B} and \texttt{Qwen3.5-27B} with an absolute Cohen's $d$~\citep{cohen2013statistical} of at least 0.5.
We then prune redundant features whose within-model correlation with a retained feature exceeds 0.8. We additionally retain two Web Shopping features because they directly characterize query behavior. We retain words per query despite its redundancy with action string length, and the indicator for repeating an identical query despite its split-half reliability of 0.47.

\paragraph{Session-level values.}
Each feature yields one value per session. Unless otherwise specified in \cref{tab:feature-def,tab:feature-def-other}, reasoning-pattern features in Web Shopping and Web Search measure the fraction of model calls containing at least one match, whereas those in Travel Planning measure matches per 1{,}000 reasoning characters.
The tables specify the units for counts, lengths, overlaps, and other features. Reasoning lengths are measured in characters of the text recorded in the trajectories. A feature is undefined when a required input is absent, and its statistic cannot be evaluated, such as mean query length in a session without search queries.

\paragraph{Alignment and aggregation.}
We align post-swap sessions by task sequence, swap point, and session index, retaining sessions available in every condition included in the comparison. If a session ends with an error in any condition, it is excluded from all
conditions. For \texttt{DeepSeek} and \texttt{Gemma}, the exclusions also include context overflows and, in Travel Planning, empty output plans.
Within each condition, we first average each feature over the retained sessions of a task sequence, omitting undefined values, and then average across task sequences with equal weight. The displayed \texttt{Qwen3.5-4B},
\texttt{Qwen3.5-27B}, and \texttt{DeepSeek} baselines use the common session cohort of the \texttt{Qwen3.5-4B/Qwen3.5-27B} and \texttt{Qwen3.5-4B/DeepSeek} comparisons. The displayed \texttt{Gemma} baseline uses the \texttt{Qwen3.5-27B/Gemma} cohort. For shifts involving \texttt{Gemma}, both no-swap history baselines are computed on the \texttt{Qwen3.5-27B/Gemma} cohort. The \texttt{Qwen3.5-27B} baseline used in these shifts may therefore differ from the \texttt{Qwen3.5-27B} baseline displayed in the table.

\paragraph{Normalized shifts.}
For recipient $Y$ and history source $X$, we compute
\[
    \pi_{X\rightarrow Y}
    = \frac{v_{X\rightarrow Y}-v_Y}{v_X-v_Y},
\]
where $v_{X\rightarrow Y}$ is the recipient's feature value under inherited history, and $v_X$ and $v_Y$ are the no-swap baselines on the corresponding session cohort. To avoid unstable normalization when the no-swap baselines are too similar, we report a shift only when $|v_X-v_Y|\geq 0.25s$, where $s$ is the standard deviation computed over the combined per-task-sequence feature values from the no-swap history conditions of all models evaluated on that cohort. We additionally require $|v_X-v_Y|>0.04$ for numerical stability. Comparisons excluded by either criterion are marked with a dash.

\label{app:behavior}
\begin{table*}[t]
    \centering
    \footnotesize
    \setlength{\tabcolsep}{4pt}
    \renewcommand{\arraystretch}{1.10}
    \begin{tabularx}{\textwidth}{@{}>{\raggedright\arraybackslash}p{0.29\textwidth}
        >{\raggedright\arraybackslash}p{0.155\textwidth}
        >{\raggedright\arraybackslash}X@{}}
        \toprule
        \textbf{Feature} & \textbf{Session-level unit} & \textbf{Definition} \\
        \midrule
        \multicolumn{3}{@{}l}{\textbf{Web Shopping}} \\
        \addlinespace[2pt]
        Words per query & Mean words & Mean number of words per \texttt{search[...]} query. \\
        Query copies option text & Fraction of queries & Queries containing at least 80\% of the content words of one listed option. \\
        Repeats identical query & 0/1 & 1 if an identical \texttt{search[...]} action is issued more than once in the session. \\
        Plan statement in reasoning & Fraction of calls & \texttt{first}, \texttt{next,}, \texttt{then}, \texttt{I will}, \texttt{I'll}, \texttt{let me}, \texttt{plan}, \texttt{step}, \texttt{now I}. \\
        Options covered by queries & Fraction of options & Listed options with at least 50\% of their content words contained in at least one query. \\
        Action string length & Mean characters & Mean character length of issued actions. \\
        Share of search actions & Fraction of actions & \texttt{search[...]} actions as a fraction of all issued actions. \\
        ASIN cited in reasoning & Fraction of calls & Calls citing a 10-character product identifier beginning with \texttt{B0}. \\
        Searches per session & Count & Number of \texttt{search[...]} actions in the session. \\
        History referenced in reasoning & Fraction of calls & \texttt{previous*}, \texttt{earlier}, \texttt{already bought/purchased/selected}, \texttt{last session/time/purchase}, \texttt{prior}, \texttt{before}, \texttt{history}. \\
        Budget terms in reasoning & Fraction of calls & \texttt{budget}, \texttt{total}, \texttt{remaining}, \texttt{spent}, \texttt{spend}, or dollar amounts. \\
        Budget arithmetic in session & 0/1 & 1 if reasoning adds dollar amounts (e.g., \texttt{\$a + \$b =}) or tracks a running total or remaining budget. \\
        \$ amounts in reasoning & Mean count per call & Mean number of dollar amounts in reasoning per model call. \\
        ``Option N'' in reasoning & Fraction of calls & \texttt{option} or \texttt{product} followed by a number. \\
        Arithmetic in reasoning & Fraction of calls & Arithmetic expressions with a result, such as \texttt{a + b =} or \texttt{= \$c}. \\
        Verification in reasoning & Fraction of calls & \texttt{verify}, \texttt{confirm*}, \texttt{check*}, \texttt{make sure}, \texttt{ensure}. \\
        Digit in query & Fraction of queries & Queries containing at least one digit. \\
        First-call reasoning length & Characters & Character length of reasoning in the first model call. \\
        Ends without purchase & 0/1 & 1 if the session ends without a \texttt{click[Buy Now]} action. \\
        List in reasoning & Fraction of calls & A line beginning with a list marker (\texttt{1.}, \texttt{1)}, \texttt{-}, \texttt{*}, or $\bullet$). \\
        Constraint check in reasoning & Fraction of calls & \texttt{compatib*}, \texttt{constraint*}, \texttt{require*}, \texttt{must}, \texttt{note(s)}, \texttt{avoid*}, \texttt{pair(s) (well) with}, \texttt{criteri*}, \texttt{instruction*}. \\
        Reaches 20-call limit & 0/1 & 1 if the session uses all 20 model calls. \\
        \bottomrule
    \end{tabularx}
    \caption{{Web Shopping feature definitions for \cref{tab:behavior-full}.}
    Units describe measurements within one session. Lexical entries summarize case-insensitive, whole-word matching patterns. An asterisk denotes any suffix, parentheses denote optional text, and slashes separate alternatives. Content words exclude stopwords and very short tokens.}
    \label{tab:feature-def}
\end{table*}

\begin{table*}[t]
    \centering
    \footnotesize
    \setlength{\tabcolsep}{4pt}
    \renewcommand{\arraystretch}{1.10}
    \begin{tabularx}{\textwidth}{@{}>{\raggedright\arraybackslash}p{0.29\textwidth}
        >{\raggedright\arraybackslash}p{0.155\textwidth}
        >{\raggedright\arraybackslash}X@{}}
        \toprule
        \textbf{Feature} & \textbf{Session-level unit} & \textbf{Definition} \\
        \midrule
        \multicolumn{3}{@{}l}{\textbf{Travel Planning}} \\
        \addlinespace[2pt]
        Budget terms in reasoning & Matches per 1k chars & \texttt{budget}, \texttt{cost(s)}, \texttt{price(s)}, dollar amounts, \texttt{per person/night}, \texttt{total}, \texttt{cheaper}, \texttt{expensive}. \\
        Meal slots filled in plan & Fraction of slots & Breakfast, lunch, and dinner slots with plan entries other than \texttt{-}. \\
        ``Format/exactly'' in reasoning & Matches per 1k chars & \texttt{format*}, \texttt{exact(ly)}. \\
        No tool call & 0/1 & 1 if the plan is produced without any tool call. \\
        Max reasoning length & Thousand characters & Maximum reasoning length across the session's model calls, divided by 1,000. \\
        Reasoning share in last call & Fraction of characters & Reasoning characters in the last model call divided by reasoning characters across all model calls. \\
        Tool calls per round & Mean count & Mean tool calls per tool-calling model call, defined only for sessions with tool calls. \\
        Co-traveller names in reasoning & Matches per 1k chars & Names of the other travelers specified in the task. \\
        History referenced in reasoning & Matches per 1k chars & \texttt{previous*}, \texttt{earlier}, \texttt{already planned/finalized/booked/confirmed}, \texttt{same as}, \texttt{join*}, \texttt{share with}, \texttt{history}, \texttt{confirmed plan}, \texttt{conversation}. \\
        \midrule
        \multicolumn{3}{@{}l}{\textbf{Web Search}} \\
        \addlinespace[2pt]
        Quoted phrase in query & Fraction of queries & Queries containing a quoted phrase. \\
        Parallel searches per turn & Fraction of tool-calling calls & Tool-calling model calls that issue more than one search simultaneously. \\
        Searches per session & Count & Number of search calls in the session. \\
        Distinct query words & Count & Number of distinct content words across all queries in the session. \\
        ``Answer now'' in reasoning & Fraction of calls & \texttt{final answer}, \texttt{exact answer}, \texttt{the answer is}, \texttt{answer:}, \texttt{provide an/my/the answer}. \\
        Overlap of consecutive queries & Mean Jaccard & Mean content-word Jaccard similarity between each query and the preceding query. \\
        ``Search again'' in reasoning & Fraction of calls & \texttt{let me search/try/look}, \texttt{search(ing) for}, \texttt{try a/another/searching/a different}, \texttt{refine}, \texttt{different query/approach/search/angle/term(s)}, \texttt{more specific}, \texttt{broader}. \\
        New top-1 document rate & Fraction of searches & Searches whose top-ranked document has not appeared as the top-ranked result of an earlier search in the session. \\
        Total reasoning length & Thousand characters & Total reasoning characters across the session's model calls, divided by 1,000. \\
        Answer starts with ``Based on'' & 0/1 & 1 if the final answer begins with ``Based on''. \\
        Hedging in reasoning & Fraction of calls & \texttt{maybe}, \texttt{perhaps}, \texttt{might}, \texttt{possibly}, \texttt{not sure}, \texttt{unclear}, \texttt{seem(s)}, \texttt{probably}, \texttt{likely}, \texttt{uncertain}. \\
        Bold ``Exact Answer'' & 0/1 & 1 if the response uses a bold ``Exact Answer'' label. \\
        Reasoning length per call & Characters & Mean number of reasoning characters per model call. \\
        \bottomrule
    \end{tabularx}
    \caption{{Travel Planning and Web Search feature definitions for \cref{tab:behavior-full}.}
    Units describe measurements within one session. Pattern notation follows
    \cref{tab:feature-def}.}
    \label{tab:feature-def-other}
\end{table*}

\clearpage

\section{Extended Experimental Results}
\label{app:exp_results}

In this section, we discuss extended single-swap experiments shown in \cref{sec:analysis}, including the results in formal reasoning (Math, Physics).

\subsection{Extended Results Under Single-Swap Setting}

\begin{figure*}[t]
    \centering
    \includegraphics[width=0.7\textwidth]{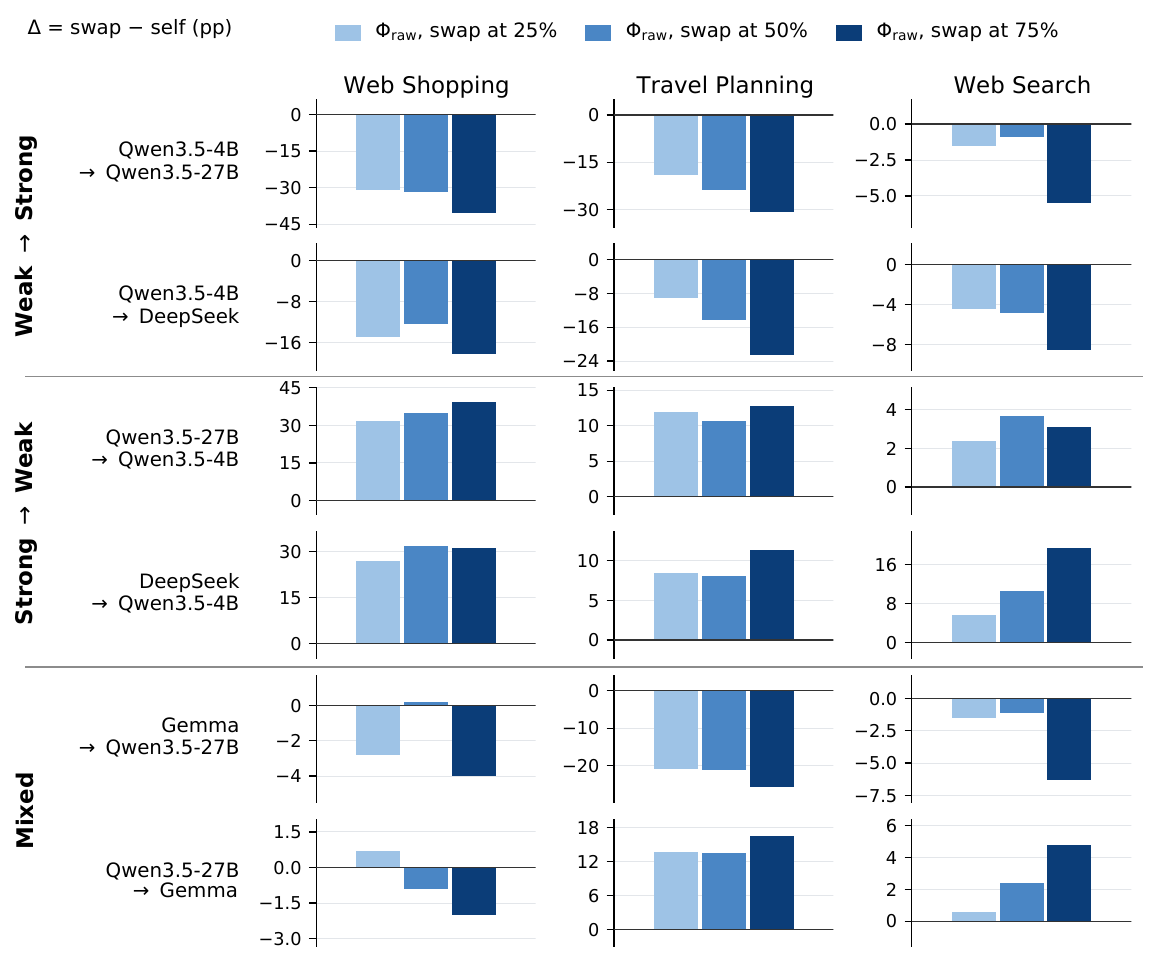}
    \caption{
        Swap position analysis of model swapping across environments. We report $\mathrm{PS}_{X\rightarrow Y} - \mathrm{PS}_{Y}$, where $X$ is the source model and $Y$ is the recipient model, using $Y$'s self-baseline for the corresponding context and subset.
    }
    \label{fig:model_swapping_position}
\end{figure*}

\paragraph{Swap point analysis.}
We examine the sensitivity of model-swapping effects to the swap point. \cref{fig:model_swapping_position} reports the change in post-swap performance, $\Delta\mathrm{PS}=\mathrm{PS}_{X\to Y}-\mathrm{PS}_{Y}$, under raw-history transfer for swaps after approximately 25\%, 50\%, or 75\% of the sessions in a task sequence.

In agentic environments, the effects remain consistent in sign across all three swap points. Swapping from \texttt{Qwen3.5-4B} to either \texttt{Qwen3.5-27B} or \texttt{DeepSeek} reduces performance, whereas the reverse swaps improve it. The directional pattern reported in the main text is therefore not confined
to a particular swap point. For both weak-to-strong pairs, the largest performance drop occurs at the 75\% swap point in each of these three environments. However, effect magnitudes do not follow a uniform monotonic trend across
all model pairs and environments.
Sign changes occur for some swaps for the \texttt{Gemma/Qwen3.5-27B} pair in Web Shopping. Since each swap point changes both the inherited history and the remaining task suffix, these comparisons do not isolate the effect of history length.

\begin{figure}[htbp!]
    \centering
    \includegraphics[width=\linewidth]{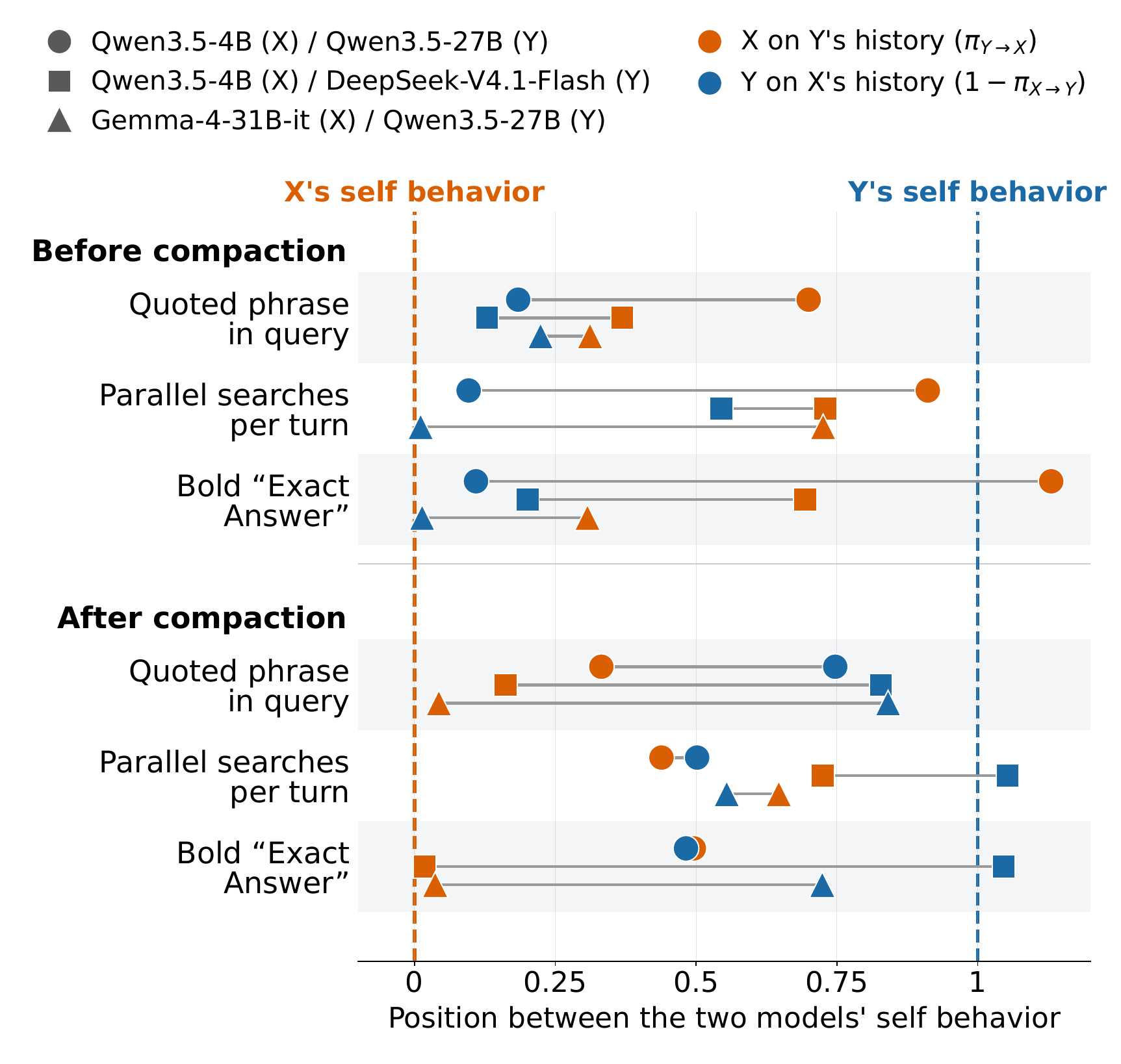}
    \caption{
    Web Search behavior transfer before and after context compaction, on the common axis of \cref{fig:behavior-transfer-main} (same features, markers, colors, and omission rule). \emph{Before compaction}: each post-swap continuation is measured from the swap until its first context compaction; the session in which that compaction occurs is cut just before it, so the recipient still sees the inherited history in full. This is the measurement shown in \cref{fig:behavior-transfer-main}. \emph{After compaction}: the same continuations measured from their first compaction onward (the remainder of the compacting session and every later session); continuations that never compact are excluded. Once compaction replaces the inherited history with a summary, most markers move back toward their own model's self behavior.}
    \label{fig:websearch-behavior-compaction}
\end{figure}

\paragraph{Behavior transfer in Web Search.}
In Web Search, retrieved documents expand the context, leading to frequent history compaction. Since our analysis focuses on behavior under raw inherited history, we examine continuations before and after the first post-swap compaction separately. \cref{fig:websearch-behavior-compaction} shows behavior transfer before compaction, when the recipient still has access to the source's raw history, consistent with the patterns observed in Web Shopping and Travel Planning. These shifts become less pronounced after compaction, consistent with the results in \cref{sec:analysis_behavior}.


\subsection{Effects of Model Swap in Formal Reasoning}
\label{app:reasoning_results}

\paragraph{Performance effect of model swap.} Overall, the directional trend observed in the \cref{sec:analysis_performance} largely persists in formal reasoning, as shown in \cref{tab:formal-reasoning-raw}. Strong-to-weak swaps generally improve the weaker recipient, while weak-to-strong swaps tend to reduce the stronger recipient's performance. This pattern also remains broadly similar after controlling for pre-swap correctness. We observe a few exceptions to this overall trend, which we examine below.

\paragraph{A beneficial weak-to-strong exception.}
Surprisingly, \texttt{DeepSeek} performs better in physics with history inherited from the weaker \texttt{Qwen3.5-4B} than with its own history, as shown in \cref{tab:formal-reasoning-raw}. This suggests that inherited context may sometimes alleviate a stronger model's existing failure pattern rather than interfere with it. Truncation at the generation-length limit offers a possible explanation for this gain. When \texttt{DeepSeek} inherits \texttt{Qwen3.5-4B}'s history, the proportion of continuation sessions truncated at the $8{,}192$-token generation cap falls from $15.2\%$ to $6.3\%$, while PS increases by $7.9$, as shown in \cref{tab:cut-ds-4b}. The gain is concentrated in cases where \texttt{DeepSeek}'s self-generated pre-swap history contains at least one truncated session, with a PS increase of $19.1$. These results suggest that inherited history may alleviate failures associated with reasoning truncation, allowing a stronger recipient to benefit from a weaker source model's history.

\begin{table}[h]
    \centering
    \small
    \setlength{\tabcolsep}{4pt}
    \begin{tabular}{@{}lrrrr@{}}
        \toprule
        & \multicolumn{2}{c}{\textbf{Physics}}
        & \multicolumn{2}{c}{\textbf{Math}} \\
        \cmidrule(lr){2-3}\cmidrule(lr){4-5}
        History used
            & Self & Inherited & Self & Inherited \\
        \midrule
        \multicolumn{5}{@{}l}{\textit{All paired swap points}} \\
        History cut ratio (\%)
            & 23.0 & 18.9 & 13.4 & 17.8 \\
        Continuation cut ratio (\%)
            & 15.2 & 6.3 & 13.9 & 14.5 \\
        Continuation PS (\%)
            & 44.9 & 52.7 & 25.3 & 24.1 \\
        \addlinespace[2pt]
        {Swap effect}
            & \multicolumn{2}{c}{${+7.9}$}
            & \multicolumn{2}{c}{${-1.3}$} \\
        \midrule
        \multicolumn{5}{@{}l}{
            \textit{Self history contains no cut sessions}
        } \\
        Continuation cut ratio (\%)
            & 15.7 & 4.4 & 14.3 & 12.2 \\
        Continuation PS (\%)
            & 61.7 & 63.4 & 28.5 & 28.1 \\
        \addlinespace[2pt]
        {Swap effect}
            & \multicolumn{2}{c}{${+1.7}$}
            & \multicolumn{2}{c}{${-0.4}$} \\
        \midrule
        \multicolumn{5}{@{}l}{
            \textit{Self history contains at least one cut session}
        } \\
        Continuation cut ratio (\%)
            & 16.7 & 9.3 & 13.4 & 17.7 \\
        Continuation PS (\%)
            & 9.5 & 28.6 & 16.6 & 15.3 \\
        \addlinespace[2pt]
        {Swap effect}
            & \multicolumn{2}{c}{${+19.1}$}
            & \multicolumn{2}{c}{${-1.3}$} \\
        \bottomrule
    \end{tabular}
    
    \caption{
        Swap effects for \texttt{DeepSeek-V4.1-Flash}. \texttt{DeepSeek} continues with either its own history (Self) or \texttt{Qwen3.5-4B}'s history (Inherited). A session is \emph{cut} when its reasoning is truncated at the 8{,}192-token maximum generation length. History cut ratio measures the percentage of cut sessions in the history available at the swap point. Continuation cut ratio measures this percentage among subsequent sessions solved by \texttt{DeepSeek} under each history. Both ratios are averaged first within and then across tasks. Swap effect is $\mathrm{PS}_{\mathrm{Inherited}} - \mathrm{PS}_{\mathrm{Self}}$. The lower blocks group paired swap points by cut occurrence in \texttt{DeepSeek}'s own history, regardless of cuts in the inherited history.
    }
    \label{tab:cut-ds-4b}
\end{table}

\begin{figure*}[t]
    \centering
    \includegraphics[width=\textwidth]{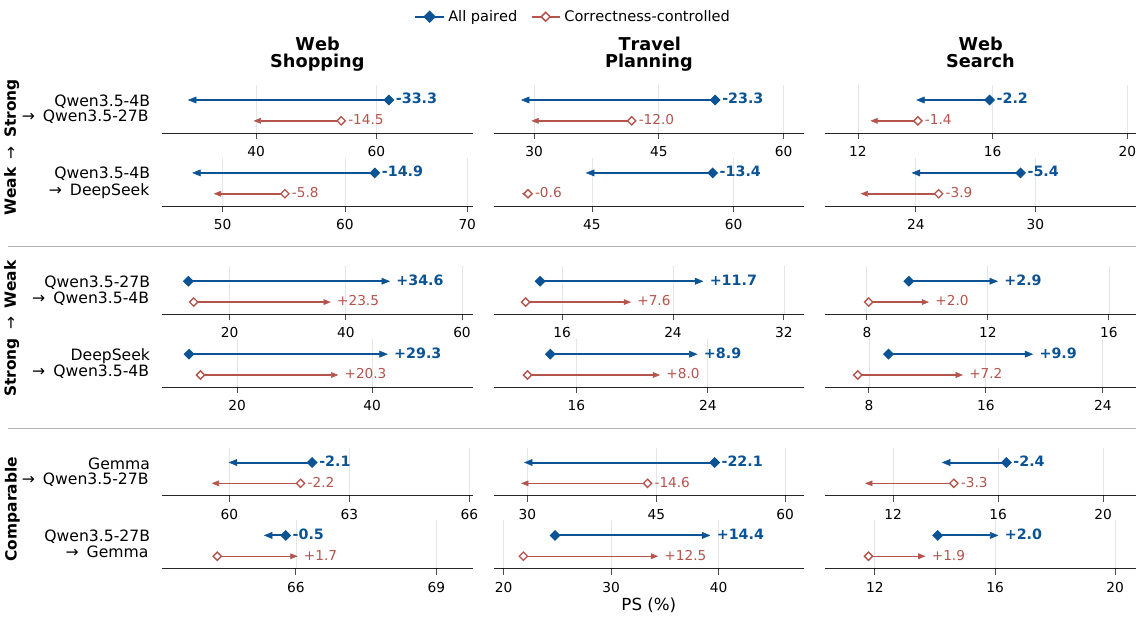}
    \caption{
        Comparison of model swapping across environments. Rows indicate the history source and receiving models, and columns indicate environments. Diamonds mark the receiving model's PS with its own history, and arrowheads mark its PS after inheriting the source model's history. Blue arrows with filled diamonds show all paired evaluations. Red arrows with hollow diamonds show the correctness-controlled subset, in which the two histories have matching correctness patterns across inherited sessions. Numerical labels report $\mathrm{PS}_{X\rightarrow Y} - \mathrm{PS}_{Y}$, where $X$ is the source model and $Y$ is the recipient model, using $Y$'s self-baseline for the corresponding context and subset.}
    \label{fig:model_swapping}
\end{figure*}

\begin{figure*}[h]
    \centering
    \includegraphics[width=\linewidth]{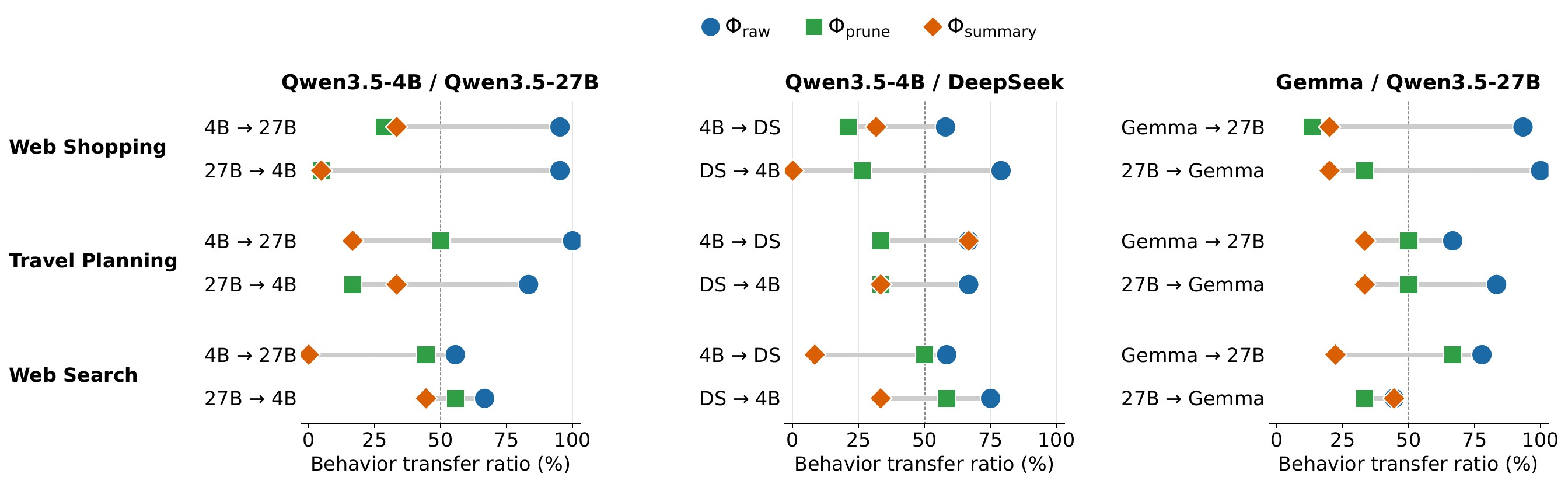}
    \caption{
    Behavior transfer is strongest under raw-history transfer, substantially reduced by reasoning pruning, and tends to be reduced further by summarization.
    The horizontal axis shows the percentage of behavioral features for which the recipient is closer to the source model's baseline than to its own ($\pi_{X\rightarrow Y}>0.5$), with both baselines measured under self-generated raw history.
    For each environment and swap direction, percentages are computed over features with valid normalized behavioral shifts.
    Feature-level results are provided in \cref{tab:behavior-full,tab:behavior-prune,tab:behavior-summary}. 4B, 27B, DS, and G denote \texttt{Qwen3.5-4B}, \texttt{Qwen3.5-27B}, \texttt{DeepSeek-V4.1-Flash}, and \texttt{Gemma-4-31B-it}, respectively.
    }
    \label{fig:median_4b_27b_extended}
\end{figure*}

\begin{table*}[htbp!]
    \centering
    \footnotesize
    \setlength{\tabcolsep}{2.5pt}
    \renewcommand{\arraystretch}{1.10}
    \begin{tabular*}{\textwidth}{@{\extracolsep{\fill}}lrrrr@{\hspace{6pt}}rr@{\hspace{6pt}}rr@{\hspace{6pt}}rr@{}}
        \toprule
        & \multicolumn{4}{c}{\textbf{Self-history baseline}}
        & \multicolumn{6}{c}{\textbf{Normalized shift}} \\
        \cmidrule(lr){2-5} \cmidrule(lr){6-11}
        Feature & 4B & 27B & DS & G
        & {\scriptsize $27\mathrm{B}\!\leftarrow\!4\mathrm{B}$}
        & {\scriptsize $4\mathrm{B}\!\leftarrow\!27\mathrm{B}$}
        & {\scriptsize $\mathrm{DS}\!\leftarrow\!4\mathrm{B}$}
        & {\scriptsize $4\mathrm{B}\!\leftarrow\!\mathrm{DS}$}
        & {\scriptsize $27\mathrm{B}\!\leftarrow\!\mathrm{G}$}
        & {\scriptsize $\mathrm{G}\!\leftarrow\!27\mathrm{B}$} \\
        \midrule
        \multicolumn{11}{@{}l}{\textbf{Web Shopping}} \\
        \addlinespace[2pt]
        Words per query & 3.82 & 8.18 & 9.14 & 10.87 & 0.85 & 0.94 & 0.73 & 0.97 & 0.88 & 0.87 \\
        Query copies option text & 0.07 & 0.39 & 0.58 & 0.58 & 0.92 & 0.87 & 0.84 & 0.90 & 1.10 & 1.07 \\
        Repeats identical query & 0.08 & 0.48 & 0.32 & 0.30 & 0.74 & 0.76 & 0.78 & 0.96 & 0.44 & 0.84 \\
        Plan statement in reasoning & 0.90 & 0.46 & 0.55 & 0.85 & 1.00 & 1.06 & 1.00 & 1.59 & 0.74 & 0.55 \\
        Options covered by queries & 0.15 & 0.48 & 0.47 & 0.52 & 0.72 & 0.93 & 0.49 & 1.04 & 1.32 & 1.21 \\
        Action string length & 23.31 & 37.81 & 40.12 & 46.79 & 0.84 & 1.12 & $-0.54$ & 0.76 & 0.74 & 0.81 \\
        Share of search actions & 0.38 & 0.50 & 0.47 & 0.52 & 0.59 & 0.69 & $-0.54$ & 0.37 & \multicolumn{1}{c}{--} & \multicolumn{1}{c}{--} \\
        ASIN cited in reasoning & 0.43 & 0.59 & 0.38 & 0.60 & 0.91 & 1.06 & 0.17 & $-1.24$ & \multicolumn{1}{c}{--} & \multicolumn{1}{c}{--} \\
        Searches per session & 2.92 & 3.77 & 3.19 & 3.85 & 0.25 & 1.12 & \multicolumn{1}{c}{--} & \multicolumn{1}{c}{--} & \multicolumn{1}{c}{--} & \multicolumn{1}{c}{--} \\
        History referenced in reasoning & 0.12 & 0.07 & 0.07 & 0.01 & 0.55 & 0.85 & 0.48 & 1.10 & 0.82 & 0.62 \\
        Budget terms in reasoning & 0.46 & 0.60 & 0.33 & 0.40 & 0.89 & 0.70 & 0.56 & 1.03 & 0.87 & 0.94 \\
        Budget arithmetic in session & 0.12 & 0.32 & 0.24 & 0.19 & 0.85 & 0.58 & 0.74 & $-0.37$ & 1.48 & 1.81 \\
        \$ amounts in reasoning & 0.97 & 1.57 & 0.65 & 0.94 & 0.92 & 0.65 & 0.53 & 1.25 & 1.01 & 0.94 \\
        ``Option N'' in reasoning & 0.05 & 0.24 & 0.11 & 0.08 & 0.94 & 0.94 & 0.45 & 0.66 & 0.95 & 0.65 \\
        Arithmetic in reasoning & 0.02 & 0.07 & 0.05 & 0.04 & 0.90 & 0.54 & 0.63 & $-0.16$ & 1.45 & 1.96 \\
        Verification in reasoning & 0.25 & 0.32 & 0.22 & 0.26 & 0.99 & 0.68 & \multicolumn{1}{c}{--} & \multicolumn{1}{c}{--} & 0.74 & 0.53 \\
        Digit in query & 0.13 & 0.42 & 0.56 & 0.63 & 0.86 & 0.96 & 0.88 & 0.96 & 0.69 & 0.73 \\
        First-call reasoning length & 1422 & 2055 & 2232 & 1203 & 0.91 & 1.62 & 0.98 & 1.29 & 0.68 & 0.54 \\
        Ends without purchase & 0.17 & 0.01 & 0.00 & 0.00 & 0.80 & 0.13 & 0.11 & 0.20 & \multicolumn{1}{c}{--} & \multicolumn{1}{c}{--} \\
        List in reasoning & 0.41 & 0.22 & 0.02 & 0.30 & 0.93 & 0.90 & 0.50 & 0.92 & 0.85 & 0.69 \\
        Constraint check in reasoning & 0.59 & 0.62 & 0.40 & 0.35 & \multicolumn{1}{c}{--} & \multicolumn{1}{c}{--} & 0.60 & 0.92 & 0.92 & 0.84 \\
        Reaches 20-call limit & 0.06 & 0.00 & 0.00 & 0.00 & 0.63 & 0.70 & 0.02 & 0.51 & \multicolumn{1}{c}{--} & \multicolumn{1}{c}{--} \\
        \midrule
        \multicolumn{11}{@{}l}{\textbf{Travel Planning}} \\
        \addlinespace[2pt]
        Budget terms in reasoning & 4.88 & 2.31 & 2.15 & 4.74 & 0.65 & 0.81 & 0.44 & 0.64 & 0.93 & 0.43 \\
        Meal slots filled in plan & 0.65 & 0.74 & 0.76 & 0.68 & 0.99 & 1.00 & 0.95 & 0.96 & 0.70 & 0.86 \\
        ``Format/exactly'' in reasoning & 1.01 & 2.53 & 0.59 & 0.46 & 0.85 & 1.02 & \multicolumn{1}{c}{--} & \multicolumn{1}{c}{--} & 1.02 & 0.67 \\
        No tool call & 0.96 & 1.00 & 0.97 & 0.98 & \multicolumn{1}{c}{--} & \multicolumn{1}{c}{--} & \multicolumn{1}{c}{--} & \multicolumn{1}{c}{--} & \multicolumn{1}{c}{--} & \multicolumn{1}{c}{--} \\
        Max reasoning length & 19.77 & 15.84 & 11.98 & 6.03 & 0.55 & 0.36 & 1.00 & $-0.07$ & $-1.79$ & 0.56 \\
        Reasoning share in last call & 0.98 & 1.00 & 0.97 & 0.93 & \multicolumn{1}{c}{--} & \multicolumn{1}{c}{--} & \multicolumn{1}{c}{--} & \multicolumn{1}{c}{--} & 0.14 & 0.96 \\
        Tool calls per round & 2.37 & 4.33 & 2.88 & 1.69 & \multicolumn{1}{c}{--} & \multicolumn{1}{c}{--} & \multicolumn{1}{c}{--} & \multicolumn{1}{c}{--} & 0.65 & 0.99 \\
        Co-traveller names in reasoning & 1.21 & 2.67 & 1.73 & 1.99 & 0.57 & 1.00 & \multicolumn{1}{c}{--} & \multicolumn{1}{c}{--} & \multicolumn{1}{c}{--} & \multicolumn{1}{c}{--} \\
        History referenced in reasoning & 0.40 & 1.00 & 0.52 & 0.70 & 0.55 & 1.00 & \multicolumn{1}{c}{--} & \multicolumn{1}{c}{--} & \multicolumn{1}{c}{--} & \multicolumn{1}{c}{--} \\
        \midrule
        \multicolumn{11}{@{}l}{\textbf{Web Search}} \\
        \addlinespace[2pt]
        Quoted phrase in query & 0.09 & 0.47 & 0.36 & 0.80 & 0.82 & 0.70 & 0.87 & 0.37 & 0.78 & 0.31 \\
        Parallel searches per turn & 0.02 & 0.14 & 0.71 & 0.04 & 0.90 & 0.91 & 0.46 & 0.73 & 0.99 & 0.73 \\
        Searches per session & 2.97 & 4.31 & 6.82 & 3.32 & 0.29 & 0.51 & 0.67 & 1.21 & 0.22 & 0.77 \\
        Distinct query words & 13.87 & 15.62 & 18.00 & 11.77 & 0.47 & 0.02 & $-0.02$ & 0.53 & 1.15 & 0.32 \\
        ``Answer now'' in reasoning & 0.14 & 0.05 & 0.13 & 0.05 & 0.35 & 0.59 & \multicolumn{1}{c}{--} & \multicolumn{1}{c}{--} & \multicolumn{1}{c}{--} & \multicolumn{1}{c}{--} \\
        Overlap of consecutive queries & 0.49 & 0.39 & 0.35 & 0.40 & 0.60 & 0.50 & 0.00 & 0.51 & \multicolumn{1}{c}{--} & \multicolumn{1}{c}{--} \\
        ``Search again'' in reasoning & 0.73 & 0.84 & 0.45 & 0.61 & 0.17 & 0.34 & 1.89 & 0.14 & 0.53 & 0.51 \\
        New top-1 document rate & 0.79 & 0.77 & 0.66 & 0.82 & \multicolumn{1}{c}{--} & \multicolumn{1}{c}{--} & 0.91 & 1.85 & $-0.01$ & 1.07 \\
        Total reasoning length & 2.56 & 2.42 & 9.75 & 3.65 & \multicolumn{1}{c}{--} & \multicolumn{1}{c}{--} & $-0.79$ & 0.58 & \multicolumn{1}{c}{--} & \multicolumn{1}{c}{--} \\
        Answer starts with ``Based on'' & 0.11 & 0.03 & 0.00 & 0.00 & 0.92 & 0.51 & 0.73 & 1.00 & \multicolumn{1}{c}{--} & \multicolumn{1}{c}{--} \\
        Hedging in reasoning & 0.11 & 0.16 & 0.57 & 0.26 & \multicolumn{1}{c}{--} & \multicolumn{1}{c}{--} & 0.62 & 0.56 & 1.19 & 0.15 \\
        Bold ``Exact Answer'' & 0.36 & 0.63 & 0.04 & 0.00 & 0.89 & 1.13 & 0.80 & 0.69 & 0.99 & 0.31 \\
        Reasoning length per call & 752 & 613 & 5613 & 1023 & \multicolumn{1}{c}{--} & \multicolumn{1}{c}{--} & 0.21 & 0.10 & 0.64 & 0.07 \\
        \bottomrule
    \end{tabular*}
    \caption{%
    Behavioral features under raw-history transfer. Baselines report continuation behavior with self-generated histories. Shift columns report $\pi_{X\rightarrow Y}$, with source $X$ and recipient $Y$. Shifts involving G use pair-specific baselines, as detailed in Appendix~\ref{app:behavior}. For Web Search, each post-swap continuation (and the corresponding self-history continuation) is measured from the swap up to its first context compaction; the compacting session is cut just before it, and later sessions are excluded.
    Dashes mark comparisons excluded by the normalization criterion; shifts are also omitted where the two models' self-history rates differ by at most 0.04, since $\pi$ is then dominated by its near-zero denominator. 4B, 27B, DS, and G denote \texttt{Qwen3.5-4B}, \texttt{Qwen3.5-27B}, \texttt{DeepSeek-V4.1-Flash}, and \texttt{Gemma-4-31B-it}, respectively.
    }
    \label{tab:behavior-full}
\end{table*}
\begin{table*}[htbp!]
    \centering
    \footnotesize
    \setlength{\tabcolsep}{2.5pt}
    \renewcommand{\arraystretch}{1.10}
    \begin{tabular*}{\textwidth}{@{\extracolsep{\fill}}lrrrr@{\hspace{6pt}}rr@{\hspace{6pt}}rr@{\hspace{6pt}}rr@{}}
        \toprule
        & \multicolumn{4}{c}{\textbf{Self-history baseline}}
        & \multicolumn{6}{c}{\textbf{Normalized shift}} \\
        \cmidrule(lr){2-5} \cmidrule(lr){6-11}
        Feature & 4B & 27B & DS & G
        & {\scriptsize $27\mathrm{B}\!\leftarrow\!4\mathrm{B}$}
        & {\scriptsize $4\mathrm{B}\!\leftarrow\!27\mathrm{B}$}
        & {\scriptsize $\mathrm{DS}\!\leftarrow\!4\mathrm{B}$}
        & {\scriptsize $4\mathrm{B}\!\leftarrow\!\mathrm{DS}$}
        & {\scriptsize $27\mathrm{B}\!\leftarrow\!\mathrm{G}$}
        & {\scriptsize $\mathrm{G}\!\leftarrow\!27\mathrm{B}$} \\
        \midrule
        \multicolumn{11}{@{}l}{\textbf{Web Shopping}} \\
        \addlinespace[2pt]
        Words per query & 3.82 & 8.18 & 9.14 & 10.87 & 0.70 & 0.44 & 0.68 & 0.61 & 0.32 & 0.76 \\
        Query copies option text & 0.07 & 0.39 & 0.58 & 0.58 & 0.83 & 0.33 & 0.81 & 0.54 & 0.50 & 0.98 \\
        Repeats identical query & 0.08 & 0.48 & 0.32 & 0.30 & 0.17 & 0.08 & 0.13 & 0.22 & 0.13 & 0.16 \\
        Plan statement in reasoning & 0.90 & 0.46 & 0.55 & 0.85 & 0.00 & 0.27 & $-0.31$ & 0.51 & 0.05 & 0.40 \\
        Options covered by queries & 0.15 & 0.48 & 0.47 & 0.52 & 0.59 & 0.20 & 0.50 & 0.50 & 0.22 & 1.22 \\
        Action string length & 23.31 & 37.81 & 40.12 & 46.79 & 0.64 & 0.21 & 0.43 & 0.40 & 0.33 & 0.75 \\
        Share of search actions & 0.38 & 0.50 & 0.47 & 0.52 & $-0.02$ & $-0.18$ & $-0.65$ & $-0.10$ & \multicolumn{1}{c}{--} & \multicolumn{1}{c}{--} \\
        ASIN cited in reasoning & 0.43 & 0.59 & 0.38 & 0.60 & 0.22 & $-0.12$ & $-0.64$ & 0.22 & \multicolumn{1}{c}{--} & \multicolumn{1}{c}{--} \\
        Searches per session & 2.92 & 3.77 & 3.19 & 3.85 & $-0.05$ & $-0.82$ & \multicolumn{1}{c}{--} & \multicolumn{1}{c}{--} & \multicolumn{1}{c}{--} & \multicolumn{1}{c}{--} \\
        History referenced in reasoning & 0.12 & 0.07 & 0.07 & 0.01 & 0.69 & $-1.79$ & 0.39 & $-1.15$ & $-0.48$ & 0.46 \\
        Budget terms in reasoning & 0.46 & 0.60 & 0.33 & 0.40 & 0.30 & $-0.03$ & $-0.31$ & 0.10 & 0.28 & 0.07 \\
        Budget arithmetic in session & 0.12 & 0.32 & 0.24 & 0.19 & 0.05 & $-0.38$ & 0.84 & $-0.59$ & 0.10 & 0.26 \\
        \$ amounts in reasoning & 0.97 & 1.57 & 0.65 & 0.94 & 0.30 & $-0.03$ & $-0.12$ & 0.45 & 0.28 & 0.07 \\
        ``Option N'' in reasoning & 0.05 & 0.24 & 0.11 & 0.08 & 0.44 & $-0.16$ & 0.24 & $-0.28$ & 0.63 & $-0.10$ \\
        Arithmetic in reasoning & 0.02 & 0.07 & 0.05 & 0.04 & 0.04 & $-0.15$ & 0.66 & $-0.37$ & 0.38 & 0.10 \\
        Verification in reasoning & 0.25 & 0.32 & 0.22 & 0.26 & $-0.48$ & 0.30 & \multicolumn{1}{c}{--} & \multicolumn{1}{c}{--} & $-0.35$ & 1.67 \\
        Digit in query & 0.13 & 0.42 & 0.56 & 0.63 & 0.81 & 0.40 & 0.79 & 0.58 & 0.27 & 0.60 \\
        First-call reasoning length & 1422 & 2055 & 2232 & 1203 & $-1.08$ & $-0.54$ & $-0.99$ & $-0.48$ & $-0.25$ & $-0.10$ \\
        Ends without purchase & 0.17 & 0.01 & 0.00 & 0.00 & 0.45 & 0.01 & 0.19 & $-0.11$ & \multicolumn{1}{c}{--} & \multicolumn{1}{c}{--} \\
        List in reasoning & 0.41 & 0.22 & 0.02 & 0.30 & 0.02 & 0.59 & 0.11 & 0.40 & 0.18 & $-0.02$ \\
        Constraint check in reasoning & 0.59 & 0.62 & 0.40 & 0.35 & \multicolumn{1}{c}{--} & \multicolumn{1}{c}{--} & 0.47 & $-0.33$ & 0.32 & 0.24 \\
        Reaches 20-call limit & 0.06 & 0.00 & 0.00 & 0.00 & 0.23 & 0.00 & 0.02 & 0.02 & \multicolumn{1}{c}{--} & \multicolumn{1}{c}{--} \\
        \midrule
        \multicolumn{11}{@{}l}{\textbf{Travel Planning}} \\
        \addlinespace[2pt]
        Budget terms in reasoning & 4.88 & 2.31 & 2.15 & 4.74 & 0.71 & $-0.54$ & $-0.25$ & $-0.51$ & 0.85 & $-0.64$ \\
        Meal slots filled in plan & 0.65 & 0.74 & 0.76 & 0.68 & 0.87 & 0.86 & 0.86 & 0.90 & 0.51 & 0.61 \\
        ``Format/exactly'' in reasoning & 1.01 & 2.53 & 0.59 & 0.46 & 1.06 & $-0.39$ & \multicolumn{1}{c}{--} & \multicolumn{1}{c}{--} & 0.63 & $-0.03$ \\
        No tool call & 0.96 & 1.00 & 0.97 & 0.98 & \multicolumn{1}{c}{--} & \multicolumn{1}{c}{--} & \multicolumn{1}{c}{--} & \multicolumn{1}{c}{--} & \multicolumn{1}{c}{--} & \multicolumn{1}{c}{--} \\
        Max reasoning length & 19.77 & 15.84 & 11.98 & 6.03 & $-0.20$ & $-2.19$ & 0.03 & $-1.47$ & $-0.10$ & 0.12 \\
        Reasoning share in last call & 0.98 & 1.00 & 0.97 & 0.93 & \multicolumn{1}{c}{--} & \multicolumn{1}{c}{--} & \multicolumn{1}{c}{--} & \multicolumn{1}{c}{--} & 0.28 & 1.00 \\
        Tool calls per round & 2.37 & 4.33 & 2.88 & 1.69 & \multicolumn{1}{c}{--} & \multicolumn{1}{c}{--} & \multicolumn{1}{c}{--} & \multicolumn{1}{c}{--} & 0.15 & 1.34 \\
        Co-traveller names in reasoning & 1.21 & 2.67 & 1.73 & 1.99 & $-0.10$ & 0.02 & \multicolumn{1}{c}{--} & \multicolumn{1}{c}{--} & \multicolumn{1}{c}{--} & \multicolumn{1}{c}{--} \\
        History referenced in reasoning & 0.40 & 1.00 & 0.52 & 0.70 & $-0.18$ & 0.31 & \multicolumn{1}{c}{--} & \multicolumn{1}{c}{--} & \multicolumn{1}{c}{--} & \multicolumn{1}{c}{--} \\
        \midrule
        \multicolumn{11}{@{}l}{\textbf{Web Search}} \\
        \addlinespace[2pt]
        Quoted phrase in query & 0.09 & 0.47 & 0.36 & 0.80 & 0.64 & 0.64 & 0.98 & 0.56 & 0.68 & 0.11 \\
        Parallel searches per turn & 0.02 & 0.14 & 0.71 & 0.04 & 0.94 & 0.88 & 0.14 & 0.87 & 0.93 & 0.74 \\
        Searches per session & 2.97 & 4.31 & 6.82 & 3.32 & $-0.07$ & 0.72 & 0.60 & 0.83 & $-0.19$ & 1.10 \\
        Distinct query words & 13.87 & 15.62 & 18.00 & 11.77 & $-0.16$ & 0.33 & $-0.13$ & 0.64 & 0.56 & 0.44 \\
        ``Answer now'' in reasoning & 0.14 & 0.05 & 0.13 & 0.05 & $-0.12$ & 0.31 & \multicolumn{1}{c}{--} & \multicolumn{1}{c}{--} & \multicolumn{1}{c}{--} & \multicolumn{1}{c}{--} \\
        Overlap of consecutive queries & 0.49 & 0.39 & 0.35 & 0.40 & 0.09 & 0.39 & $-0.10$ & 0.70 & \multicolumn{1}{c}{--} & \multicolumn{1}{c}{--} \\
        ``Search again'' in reasoning & 0.73 & 0.84 & 0.45 & 0.61 & 0.06 & $-0.14$ & 1.88 & 0.04 & 0.34 & 0.41 \\
        New top-1 document rate & 0.79 & 0.77 & 0.66 & 0.82 & \multicolumn{1}{c}{--} & \multicolumn{1}{c}{--} & 1.01 & 1.13 & $-0.28$ & 1.34 \\
        Total reasoning length & 2.56 & 2.42 & 9.75 & 3.65 & \multicolumn{1}{c}{--} & \multicolumn{1}{c}{--} & 0.21 & 0.11 & \multicolumn{1}{c}{--} & \multicolumn{1}{c}{--} \\
        Answer starts with ``Based on'' & 0.11 & 0.03 & 0.00 & 0.00 & 0.86 & 0.78 & 0.00 & $-0.10$ & \multicolumn{1}{c}{--} & \multicolumn{1}{c}{--} \\
        Hedging in reasoning & 0.11 & 0.16 & 0.57 & 0.26 & \multicolumn{1}{c}{--} & \multicolumn{1}{c}{--} & 0.32 & $-0.06$ & 0.51 & $-0.66$ \\
        Bold ``Exact Answer'' & 0.36 & 0.63 & 0.04 & 0.00 & 0.67 & 1.08 & 0.85 & 0.63 & 0.93 & 0.12 \\
        Reasoning length per call & 752 & 613 & 5613 & 1023 & \multicolumn{1}{c}{--} & \multicolumn{1}{c}{--} & 0.63 & 0.06 & 0.85 & $-0.72$ \\
        \bottomrule
    \end{tabular*}
    \caption{%
    Behavioral features under swap-time reasoning pruning ($\Phi_{\text{prune}}$). Baselines report continuation behavior with self-generated histories, as in Table~\ref{tab:behavior-full}. Shift columns report $\pi_{X\rightarrow Y}$ for recipient $Y$ continuing the history of source $X$ transferred by $\Phi_{\text{prune}}$, computed on the same sessions as the transferred-history condition. Dashes mark comparisons excluded by the normalization criterion; shifts are also omitted where the two models' self-history rates differ by at most 0.04, since $\pi$ is then dominated by its near-zero denominator. 4B, 27B, DS, and G denote \texttt{Qwen3.5-4B}, \texttt{Qwen3.5-27B}, \texttt{DeepSeek-V4.1-Flash}, and \texttt{Gemma-4-31B-it}, respectively. For Web Search, each post-swap continuation (and the corresponding self-history continuation) is measured from the swap up to its first context compaction; the compacting session is cut just before it, and later sessions are excluded.
    }
    \label{tab:behavior-prune}
\end{table*}
\begin{table*}[htbp!]
    \centering
    \footnotesize
    \setlength{\tabcolsep}{2.5pt}
    \renewcommand{\arraystretch}{1.10}
    \begin{tabular*}{\textwidth}{@{\extracolsep{\fill}}lrrrr@{\hspace{6pt}}rr@{\hspace{6pt}}rr@{\hspace{6pt}}rr@{}}
        \toprule
        & \multicolumn{4}{c}{\textbf{Self-history baseline}}
        & \multicolumn{6}{c}{\textbf{Normalized shift}} \\
        \cmidrule(lr){2-5} \cmidrule(lr){6-11}
        Feature & 4B & 27B & DS & G
        & {\scriptsize $27\mathrm{B}\!\leftarrow\!4\mathrm{B}$}
        & {\scriptsize $4\mathrm{B}\!\leftarrow\!27\mathrm{B}$}
        & {\scriptsize $\mathrm{DS}\!\leftarrow\!4\mathrm{B}$}
        & {\scriptsize $4\mathrm{B}\!\leftarrow\!\mathrm{DS}$}
        & {\scriptsize $27\mathrm{B}\!\leftarrow\!\mathrm{G}$}
        & {\scriptsize $\mathrm{G}\!\leftarrow\!27\mathrm{B}$} \\
        \midrule
        \multicolumn{11}{@{}l}{\textbf{Web Shopping}} \\
        \addlinespace[2pt]
        Words per query & 3.82 & 8.18 & 9.14 & 10.87 & 0.46 & $-0.17$ & 0.40 & $-0.08$ & 0.04 & 0.00 \\
        Query copies option text & 0.07 & 0.39 & 0.58 & 0.58 & 0.53 & $-0.16$ & 0.52 & $-0.06$ & 0.00 & $-0.11$ \\
        Repeats identical query & 0.08 & 0.48 & 0.32 & 0.30 & 0.30 & 0.01 & 0.33 & 0.02 & $-0.06$ & 0.56 \\
        Plan statement in reasoning & 0.90 & 0.46 & 0.55 & 0.85 & 0.01 & 0.07 & 0.78 & 0.09 & 0.03 & 0.09 \\
        Options covered by queries & 0.15 & 0.48 & 0.47 & 0.52 & 0.43 & $-0.17$ & 0.13 & $-0.08$ & 0.39 & 0.12 \\
        Action string length & 23.31 & 37.81 & 40.12 & 46.79 & 0.43 & $-0.21$ & $-3.82$ & $-0.13$ & 0.35 & $-0.83$ \\
        Share of search actions & 0.38 & 0.50 & 0.47 & 0.52 & 0.23 & $-0.62$ & $-0.74$ & $-0.82$ & \multicolumn{1}{c}{--} & \multicolumn{1}{c}{--} \\
        ASIN cited in reasoning & 0.43 & 0.59 & 0.38 & 0.60 & 0.40 & 0.36 & 0.52 & $-0.99$ & \multicolumn{1}{c}{--} & \multicolumn{1}{c}{--} \\
        Searches per session & 2.92 & 3.77 & 3.19 & 3.85 & 0.20 & $-0.41$ & \multicolumn{1}{c}{--} & \multicolumn{1}{c}{--} & \multicolumn{1}{c}{--} & \multicolumn{1}{c}{--} \\
        History referenced in reasoning & 0.12 & 0.07 & 0.07 & 0.01 & 0.96 & $-0.43$ & 0.52 & $-0.31$ & $-0.60$ & 0.26 \\
        Budget terms in reasoning & 0.46 & 0.60 & 0.33 & 0.40 & 0.19 & $-0.02$ & 0.13 & $-0.10$ & 0.16 & 0.24 \\
        Budget arithmetic in session & 0.12 & 0.32 & 0.24 & 0.19 & $-0.01$ & $-0.08$ & 1.35 & $-0.07$ & $-0.51$ & 2.25 \\
        \$ amounts in reasoning & 0.97 & 1.57 & 0.65 & 0.94 & 0.02 & 0.02 & $-0.06$ & $-0.22$ & $-0.19$ & 0.45 \\
        ``Option N'' in reasoning & 0.05 & 0.24 & 0.11 & 0.08 & 0.56 & 0.08 & $-0.48$ & 0.34 & 0.58 & $-0.04$ \\
        Arithmetic in reasoning & 0.02 & 0.07 & 0.05 & 0.04 & $-0.08$ & $-0.17$ & 1.00 & $-0.16$ & $-1.18$ & 1.28 \\
        Verification in reasoning & 0.25 & 0.32 & 0.22 & 0.26 & $-0.89$ & 0.62 & \multicolumn{1}{c}{--} & \multicolumn{1}{c}{--} & $-0.88$ & 0.87 \\
        Digit in query & 0.13 & 0.42 & 0.56 & 0.63 & 0.63 & $-0.20$ & 0.60 & $-0.07$ & $-0.36$ & 0.38 \\
        First-call reasoning length & 1422 & 2055 & 2232 & 1203 & 1.13 & $-0.64$ & 0.37 & $-0.51$ & 0.63 & 0.31 \\
        Ends without purchase & 0.17 & 0.01 & 0.00 & 0.00 & 0.64 & $-0.73$ & 0.13 & $-0.88$ & \multicolumn{1}{c}{--} & \multicolumn{1}{c}{--} \\
        List in reasoning & 0.41 & 0.22 & 0.02 & 0.30 & 1.05 & 0.02 & 0.31 & 0.01 & 2.84 & 0.06 \\
        Constraint check in reasoning & 0.59 & 0.62 & 0.40 & 0.35 & \multicolumn{1}{c}{--} & \multicolumn{1}{c}{--} & 0.48 & 0.14 & 0.31 & 0.29 \\
        Reaches 20-call limit & 0.06 & 0.00 & 0.00 & 0.00 & 0.37 & $-2.15$ & 0.15 & $-1.83$ & \multicolumn{1}{c}{--} & \multicolumn{1}{c}{--} \\
        \midrule
        \multicolumn{11}{@{}l}{\textbf{Travel Planning}} \\
        \addlinespace[2pt]
        Budget terms in reasoning & 4.88 & 2.31 & 2.15 & 4.74 & 1.42 & $-1.30$ & 0.55 & $-1.55$ & 1.47 & $-0.67$ \\
        Meal slots filled in plan & 0.65 & 0.74 & 0.76 & 0.68 & 0.37 & 1.03 & 0.40 & 0.92 & 0.19 & 0.76 \\
        ``Format/exactly'' in reasoning & 1.01 & 2.53 & 0.59 & 0.46 & 0.30 & 0.29 & \multicolumn{1}{c}{--} & \multicolumn{1}{c}{--} & 0.03 & 0.00 \\
        No tool call & 0.96 & 1.00 & 0.97 & 0.98 & \multicolumn{1}{c}{--} & \multicolumn{1}{c}{--} & \multicolumn{1}{c}{--} & \multicolumn{1}{c}{--} & \multicolumn{1}{c}{--} & \multicolumn{1}{c}{--} \\
        Max reasoning length & 19.77 & 15.84 & 11.98 & 6.03 & $-0.07$ & $-1.51$ & 1.91 & $-0.66$ & 0.05 & 0.20 \\
        Reasoning share in last call & 0.98 & 1.00 & 0.97 & 0.93 & \multicolumn{1}{c}{--} & \multicolumn{1}{c}{--} & \multicolumn{1}{c}{--} & \multicolumn{1}{c}{--} & 1.45 & $-2.82$ \\
        Tool calls per round & 2.37 & 4.33 & 2.88 & 1.69 & \multicolumn{1}{c}{--} & \multicolumn{1}{c}{--} & \multicolumn{1}{c}{--} & \multicolumn{1}{c}{--} & 0.47 & 0.72 \\
        Co-traveller names in reasoning & 1.21 & 2.67 & 1.73 & 1.99 & 0.17 & 0.37 & \multicolumn{1}{c}{--} & \multicolumn{1}{c}{--} & \multicolumn{1}{c}{--} & \multicolumn{1}{c}{--} \\
        History referenced in reasoning & 0.40 & 1.00 & 0.52 & 0.70 & 0.28 & 0.59 & \multicolumn{1}{c}{--} & \multicolumn{1}{c}{--} & \multicolumn{1}{c}{--} & \multicolumn{1}{c}{--} \\
        \midrule
        \multicolumn{11}{@{}l}{\textbf{Web Search}} \\
        \addlinespace[2pt]
        Quoted phrase in query & 0.09 & 0.47 & 0.36 & 0.80 & 0.44 & 0.07 & $-0.08$ & 0.17 & $-0.38$ & 0.08 \\
        Parallel searches per turn & 0.02 & 0.14 & 0.71 & 0.04 & $-2.45$ & 2.90 & $-0.07$ & 0.18 & $-3.30$ & 2.45 \\
        Searches per session & 2.97 & 4.31 & 6.82 & 3.32 & $-3.93$ & 4.04 & $-2.40$ & 1.30 & $-5.43$ & 6.31 \\
        Distinct query words & 13.87 & 15.62 & 18.00 & 11.77 & $-3.80$ & 3.53 & $-2.77$ & 1.21 & $-1.58$ & 1.51 \\
        ``Answer now'' in reasoning & 0.14 & 0.05 & 0.13 & 0.05 & 0.34 & 0.40 & \multicolumn{1}{c}{--} & \multicolumn{1}{c}{--} & \multicolumn{1}{c}{--} & \multicolumn{1}{c}{--} \\
        Overlap of consecutive queries & 0.49 & 0.39 & 0.35 & 0.40 & $-0.92$ & 0.87 & $-0.28$ & 0.33 & \multicolumn{1}{c}{--} & \multicolumn{1}{c}{--} \\
        ``Search again'' in reasoning & 0.73 & 0.84 & 0.45 & 0.61 & 0.49 & $-0.16$ & 0.22 & $-0.26$ & 0.20 & 0.31 \\
        New top-1 document rate & 0.79 & 0.77 & 0.66 & 0.82 & \multicolumn{1}{c}{--} & \multicolumn{1}{c}{--} & $-0.79$ & 1.77 & $-2.25$ & 3.25 \\
        Total reasoning length & 2.56 & 2.42 & 9.75 & 3.65 & \multicolumn{1}{c}{--} & \multicolumn{1}{c}{--} & $-1.07$ & 0.61 & \multicolumn{1}{c}{--} & \multicolumn{1}{c}{--} \\
        Answer starts with ``Based on'' & 0.11 & 0.03 & 0.00 & 0.00 & $-0.07$ & 0.20 & 0.00 & $-0.06$ & \multicolumn{1}{c}{--} & \multicolumn{1}{c}{--} \\
        Hedging in reasoning & 0.11 & 0.16 & 0.57 & 0.26 & \multicolumn{1}{c}{--} & \multicolumn{1}{c}{--} & $-0.24$ & 0.12 & 0.80 & $-0.42$ \\
        Bold ``Exact Answer'' & 0.36 & 0.63 & 0.04 & 0.00 & $-0.07$ & 0.30 & $-0.06$ & 0.36 & 0.03 & 0.00 \\
        Reasoning length per call & 752 & 613 & 5613 & 1023 & \multicolumn{1}{c}{--} & \multicolumn{1}{c}{--} & 0.60 & 0.03 & 0.50 & $-0.67$ \\
        \bottomrule
    \end{tabular*}
    \caption{%
    Behavioral features under session-wise summarization ($\Phi_{\text{summary}}$). Baselines report continuation behavior with self-generated histories, as in \cref{tab:behavior-full}. Shift columns report $\pi_{X\rightarrow Y}$ for recipient $Y$ continuing the history of source $X$ transferred by $\Phi_{\text{summary}}$, computed on the same sessions as the transferred-history condition. Dashes mark comparisons excluded by the normalization criterion; shifts are also omitted where the two models' self-history rates differ by at most 0.04, since $\pi$ is then dominated by its near-zero denominator. 4B, 27B, DS, and G denote \texttt{Qwen3.5-4B}, \texttt{Qwen3.5-27B}, \texttt{DeepSeek-V4.1-Flash}, and \texttt{Gemma-4-31B-it}, respectively. For Web Search, each post-swap continuation (and the corresponding self-history continuation) is measured from the swap up to its first context compaction; the compacting session is cut just before it, and later sessions are excluded. Under $\Phi_{\text{summary}}$, both recipients in a pair search more, and more often in parallel, than either model does on its own history, so the search-volume features take large shifts of opposite sign in the two directions; these reflect the summarized context itself rather than transfer from the source model.
    }
    \label{tab:behavior-summary}
\end{table*}

\begin{table*}[t]
    \centering
    \small
    \setlength{\tabcolsep}{4.5pt}
    \renewcommand{\arraystretch}{1.08}
    \begin{tabular}{@{}lllrr@{}}
        \toprule
        \textbf{Direction}
        & \textbf{Model / Swap}
        & \textbf{Subset}
        & \textbf{Math}
        & \textbf{Physics} \\
        \midrule

        \multirow{4}{*}{\shortstack[l]{No swap\\(Self)}}
        & \texttt{Qwen3.5-4B}
        &  & 14.8 & 25.6 \\
        & \texttt{Qwen3.5-27B}
        &  & 25.2 & 55.4 \\
        & \texttt{DeepSeek}
        &  & 25.3 & 44.9 \\
        & \texttt{Gemma}
        &  & 28.4 & 57.4 \\
        \midrule

        \multirow{4}{*}{\shortstack[l]{Weak\\$\rightarrow$ strong}}
        & \multirow{2}{*}{\shortstack[l]{
            \texttt{Qwen3.5-4B}\\
            $\rightarrow$ \texttt{Qwen3.5-27B}
        }}
        & All paired
        & 26.4\,{\scriptsize$(+1.2)$}
        & 47.6\,{\scriptsize$(-7.8)$} \\
        & & Correctness
        & 26.1\,{\scriptsize$(+1.0)$}
        & 59.5\,{\scriptsize$(-2.4)$} \\
        \cmidrule(lr){2-5}

        & \multirow{2}{*}{\shortstack[l]{
            \texttt{Qwen3.5-4B}\\
            $\rightarrow$ \texttt{DeepSeek}
        }}
        & All paired
        & 24.1\,{\scriptsize$(-1.2)$}
        & 52.7\,{\scriptsize$(+7.8)$} \\
        & & Correctness
        & 28.9\,{\scriptsize$(-0.7)$}
        & 68.2\,{\scriptsize$(+6.1)$} \\
        \midrule

        \multirow{4}{*}{\shortstack[l]{Strong\\$\rightarrow$ weak}}
        & \multirow{2}{*}{\shortstack[l]{
            \texttt{Qwen3.5-27B}\\
            $\rightarrow$ \texttt{Qwen3.5-4B}
        }}
        & All paired
        & 16.9\,{\scriptsize$(+2.1)$}
        & 39.3\,{\scriptsize$(+13.7)$} \\
        & & Correctness
        & 13.2\,{\scriptsize$(-0.9)$}
        & 45.2\,{\scriptsize$(+9.5)$} \\
        \cmidrule(lr){2-5}

        & \multirow{2}{*}{\shortstack[l]{
            \texttt{DeepSeek}\\
            $\rightarrow$ \texttt{Qwen3.5-4B}
        }}
        & All paired
        & 17.1\,{\scriptsize$(+2.3)$}
        & 38.7\,{\scriptsize$(+13.1)$} \\
        & & Correctness
        & 20.0\,{\scriptsize$(+2.1)$}
        & 51.5\,{\scriptsize$(+9.1)$} \\
        \midrule

        \multirow{4}{*}{Mixed}
        & \multirow{2}{*}{\shortstack[l]{
            \texttt{Gemma}\\
            $\rightarrow$ \texttt{Qwen3.5-27B}
        }}
        & All paired
        & 26.3\,{\scriptsize$(+1.1)$}
        & 58.9\,{\scriptsize$(+3.5)$} \\
        & & Correctness
        & 28.3\,{\scriptsize$(+3.1)$}
        & 67.3\,{\scriptsize$(+7.7)$} \\
        \cmidrule(lr){2-5}

        & \multirow{2}{*}{\shortstack[l]{
            \texttt{Qwen3.5-27B}\\
            $\rightarrow$ \texttt{Gemma}
        }}
        & All paired
        & 27.8\,{\scriptsize$(-0.6)$}
        & 54.0\,{\scriptsize$(-3.4)$} \\
        & & Correctness
        & 33.0\,{\scriptsize$(-1.0)$}
        & 55.4\,{\scriptsize$(-12.1)$} \\
        \bottomrule
    \end{tabular}

    \caption{
        Post-swap performance under raw-history transfer on the formal reasoning environments. Self rows report performance without model swapping. For each directed swap, we report all paired evaluations and the correctness-controlled subset (Correctness), in which the source and recipient histories have matching correctness patterns across all pre-swap sessions. Values in parentheses report $\mathrm{PS}_{X\rightarrow Y} - \mathrm{PS}_{Y}$, where $X$ is the source model and $Y$ is the recipient model, using $Y$'s self baseline for the corresponding context and subset.
    }
    \label{tab:formal-reasoning-raw}
\end{table*}

\begin{figure*}[t]
    \centering
    \includegraphics[width=\linewidth]{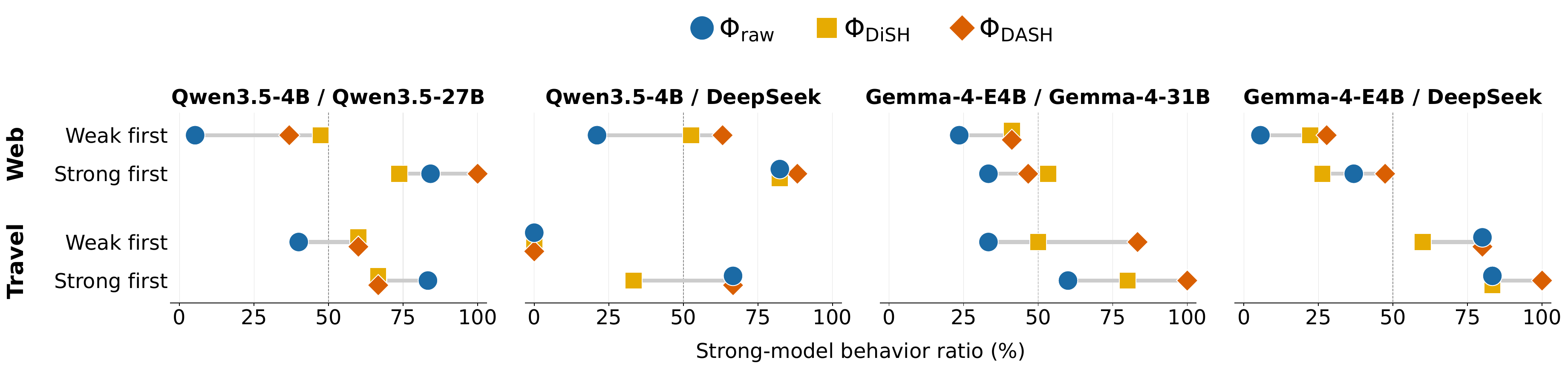}
    \caption{
    Direction-aware summarization shifts post-swap behavior toward the strong model, especially under weak-first routing. The horizontal axis shows the percentage of behavioral features whose post-swap means are closer to the strong model's baseline than to the weak model's ($(v_{\mathrm{post}}-v_W)/(v_S-v_W)>0.5$), with both baselines measured under self-generated raw history. Feature means pool sessions after the first swap regardless of the acting model, and percentages are computed over features with valid normalized shifts.
    }
    \label{fig:multiple_swap_full}
\end{figure*}

\clearpage

\section{Tabulated Results}
We provide tabulated results for figures in the main text in \cref{tab:memoryarena-model-swapping,tab:memoryarena-swap-controlled,tab:multiswap}

\begin{table*}[t]
    \centering
    \small
    \setlength{\tabcolsep}{5pt}
    \renewcommand{\arraystretch}{1.05}
    \begin{tabular}{@{\extracolsep{\fill}}lllrrr@{}}
        \toprule
        \textbf{Swap}
        & \textbf{Model / Swap}
        & \textbf{Context}
        & \textbf{Web Shopping}
        & \textbf{Travel Planning}
        & \textbf{Web Search} \\
        \midrule

        \multirow{4}{*}{\shortstack[l]{No swap\\(Self)}}
        & \texttt{Qwen3.5-4B}
        & $\Phi_{\mathrm{raw}}$
        & 12.9 & 14.4 & 9.4 \\
        \cmidrule(lr){2-6}

        & \texttt{Qwen3.5-27B}
        & $\Phi_{\mathrm{raw}}$
        & 62.1 & 51.8 & 15.9 \\
        \cmidrule(lr){2-6}

        & \texttt{Gemma}
        & $\Phi_{\mathrm{raw}}$
        & 65.8 & 24.8 & 14.1 \\
        \cmidrule(lr){2-6}

        & \texttt{DeepSeek}
        & $\Phi_{\mathrm{raw}}$
        & 62.4 & 57.8 & 29.3 \\
        \midrule

        \multirow{6}{*}{\shortstack[l]{Weak\\$\rightarrow$ strong}}
        & \multirow{3}{*}{\shortstack[l]{
            \texttt{Qwen3.5-4B}\\
            $\rightarrow$ \texttt{Qwen3.5-27B}
        }}
        & $\Phi_{\mathrm{raw}}$
        & 28.7\,{\scriptsize$(-33.4)$}
        & 28.5\,{\scriptsize$(-23.3)$}
        & 13.8\,{\scriptsize$(-2.1)$} \\
        & & $\Phi_{\mathrm{prune}}$
        & 35.9\,{\scriptsize$(-26.2)$}
        & 26.2\,{\scriptsize$(-25.6)$}
        & 13.7\,{\scriptsize$(-2.2)$} \\
        & & $\Phi_{\mathrm{summary}}$
        & 37.0\,{\scriptsize$(-25.1)$}
        & 37.9\,{\scriptsize$(-13.9)$}
        & 14.7\,{\scriptsize$(-1.2)$} \\
        \cmidrule(lr){2-6}

        & \multirow{3}{*}{\shortstack[l]{
            \texttt{Qwen3.5-4B}\\
            $\rightarrow$ \texttt{DeepSeek}
        }}
        & $\Phi_{\mathrm{raw}}$
        & 47.6\,{\scriptsize$(-14.8)$}
        & 44.4\,{\scriptsize$(-13.4)$}
        & 23.8\,{\scriptsize$(-5.5)$} \\
        & & $\Phi_{\mathrm{prune}}$
        & 46.3\,{\scriptsize$(-16.1)$}
        & 41.7\,{\scriptsize$(-16.1)$}
        & 26.5\,{\scriptsize$(-2.8)$} \\
        & & $\Phi_{\mathrm{summary}}$
        & 51.0\,{\scriptsize$(-11.4)$}
        & 53.7\,{\scriptsize$(-4.1)$}
        & 26.7\,{\scriptsize$(-2.6)$} \\
        \midrule

        \multirow{6}{*}{\shortstack[l]{Strong\\$\rightarrow$ weak}}
        & \multirow{3}{*}{\shortstack[l]{
            \texttt{Qwen3.5-27B}\\
            $\rightarrow$ \texttt{Qwen3.5-4B}
        }}
        & $\Phi_{\mathrm{raw}}$
        & 47.5\,{\scriptsize$(+34.6)$}
        & 26.1\,{\scriptsize$(+11.7)$}
        & 12.3\,{\scriptsize$(+2.9)$} \\
        & & $\Phi_{\mathrm{prune}}$
        & 17.3\,{\scriptsize$(+4.4)$}
        & 8.4\,{\scriptsize$(-6.0)$}
        & 11.7\,{\scriptsize$(+2.3)$} \\
        & & $\Phi_{\mathrm{summary}}$
        & 10.7\,{\scriptsize$(-2.2)$}
        & 21.7\,{\scriptsize$(+7.3)$}
        & 11.7\,{\scriptsize$(+2.3)$} \\
        \cmidrule(lr){2-6}

        & \multirow{3}{*}{\shortstack[l]{
            \texttt{DeepSeek}\\
            $\rightarrow$ \texttt{Qwen3.5-4B}
        }}
        & $\Phi_{\mathrm{raw}}$
        & 42.2\,{\scriptsize$(+29.3)$}
        & 23.3\,{\scriptsize$(+8.9)$}
        & 19.2\,{\scriptsize$(+9.8)$} \\
        & & $\Phi_{\mathrm{prune}}$
        & 17.1\,{\scriptsize$(+4.2)$}
        & 8.4\,{\scriptsize$(-6.0)$}
        & 19.7\,{\scriptsize$(+10.3)$} \\
        & & $\Phi_{\mathrm{summary}}$
        & 10.7\,{\scriptsize$(-2.2)$}
        & 21.6\,{\scriptsize$(+7.2)$}
        & 18.5\,{\scriptsize$(+9.1)$} \\
        \midrule

        \multirow{6}{*}{Mixed}
        & \multirow{3}{*}{\shortstack[l]{
            \texttt{Gemma}\\
            $\rightarrow$ \texttt{Qwen3.5-27B}
        }}
        & $\Phi_{\mathrm{raw}}$
        & 60.0\,{\scriptsize$(-2.1)$}
        & 29.7\,{\scriptsize$(-22.1)$}
        & 13.9\,{\scriptsize$(-2.0)$} \\
        & & $\Phi_{\mathrm{prune}}$
        & 55.1\,{\scriptsize$(-7.0)$}
        & 32.7\,{\scriptsize$(-19.1)$}
        & 14.6\,{\scriptsize$(-1.3)$} \\
        & & $\Phi_{\mathrm{summary}}$
        & 53.7\,{\scriptsize$(-8.4)$}
        & 51.6\,{\scriptsize$(-0.2)$}
        & 15.0\,{\scriptsize$(-0.9)$} \\
        \cmidrule(lr){2-6}

        & \multirow{3}{*}{\shortstack[l]{
            \texttt{Qwen3.5-27B}\\
            $\rightarrow$ \texttt{Gemma}
        }}
        & $\Phi_{\mathrm{raw}}$
        & 65.3\,{\scriptsize$(-0.5)$}
        & 39.2\,{\scriptsize$(+14.4)$}
        & 16.1\,{\scriptsize$(+2.0)$} \\
        & & $\Phi_{\mathrm{prune}}$
        & 67.0\,{\scriptsize$(+1.2)$}
        & 35.4\,{\scriptsize$(+10.6)$}
        & 16.7\,{\scriptsize$(+2.6)$} \\
        & & $\Phi_{\mathrm{summary}}$
        & 66.5\,{\scriptsize$(+0.7)$}
        & 39.9\,{\scriptsize$(+15.1)$}
        & 16.8\,{\scriptsize$(+2.7)$} \\
        \bottomrule
    \end{tabular}
    \caption{
        Post-swap performance on the agentic environments in MemoryArena. $\mathcal{M}_A \rightarrow \mathcal{M}_B$ denotes a swap from source model $\mathcal{M}_A$ to recipient model $\mathcal{M}_B$. Values in parentheses report $\mathrm{PS}_{\mathrm{swap}}-\mathrm{PS}_{\mathrm{self}}$. For swap-time pruning and summarization, subsequent sessions retain raw history.
        All summaries are written by \texttt{Qwen3.5-27B}. \texttt{DeepSeek} and \texttt{Gemma} denote \texttt{DeepSeek-V4.1-Flash} and \texttt{Gemma-4-31B-it}, respectively.
    }
    \label{tab:memoryarena-model-swapping}
\end{table*}

\begin{table*}[t]
    \centering
    \small
    \setlength{\tabcolsep}{5pt}
    \renewcommand{\arraystretch}{1.08}
    \begin{tabular}{@{\extracolsep{\fill}}lllrrr@{}}
        \toprule
        \textbf{Swap}
        & \textbf{Model / Swap}
        & \textbf{Subset}
        & \textbf{Web Shopping}
        & \textbf{Travel Planning}
        & \textbf{Web Search} \\
        \midrule

        \multirow{6}{*}{\shortstack[l]{Weak\\$\rightarrow$ strong}}
        & \multirow{3}{*}{\shortstack[l]{
            \texttt{Qwen3.5-4B}\\
            $\rightarrow$ \texttt{Qwen3.5-27B}
        }}
        & All paired
        & 28.7\,{\scriptsize$(-33.4)$}
        & 28.5\,{\scriptsize$(-23.3)$}
        & 13.8\,{\scriptsize$(-2.1)$} \\
        & & Correctness-controlled
        & 39.7\,{\scriptsize$(-14.5)$}
        & 29.8\,{\scriptsize$(-12.0)$}
        & 12.4\,{\scriptsize$(-1.4)$} \\
        & & {\scriptsize \# tasks (controlled / all)}
        & {\scriptsize 45 / 150}
        & {\scriptsize 111 / 270}
        & {\scriptsize 204 / 221} \\
        \cmidrule(lr){2-6}

        & \multirow{3}{*}{\shortstack[l]{
            \texttt{Qwen3.5-4B}\\
            $\rightarrow$ \texttt{DeepSeek}
        }}
        & All paired
        & 47.6\,{\scriptsize$(-14.8)$}
        & 44.4\,{\scriptsize$(-13.4)$}
        & 23.8\,{\scriptsize$(-5.5)$} \\
        & & Correctness-controlled
        & 49.3\,{\scriptsize$(-5.8)$}
        & 37.6\,{\scriptsize$(-0.6)$}
        & 21.3\,{\scriptsize$(-3.9)$} \\
        & & {\scriptsize \# tasks (controlled / all)}
        & {\scriptsize 36 / 150}
        & {\scriptsize 88 / 270}
        & {\scriptsize 189 / 221} \\
        \midrule

        \multirow{6}{*}{\shortstack[l]{Strong\\$\rightarrow$ weak}}
        & \multirow{3}{*}{\shortstack[l]{
            \texttt{Qwen3.5-27B}\\
            $\rightarrow$ \texttt{Qwen3.5-4B}
        }}
        & All paired
        & 47.5\,{\scriptsize$(+34.6)$}
        & 26.1\,{\scriptsize$(+11.7)$}
        & 12.3\,{\scriptsize$(+2.9)$} \\
        & & Correctness-controlled
        & 37.3\,{\scriptsize$(+23.5)$}
        & 20.9\,{\scriptsize$(+7.6)$}
        & 10.0\,{\scriptsize$(+2.0)$} \\
        & & {\scriptsize \# tasks (controlled / all)}
        & {\scriptsize 45 / 150}
        & {\scriptsize 111 / 270}
        & {\scriptsize 204 / 221} \\
        \cmidrule(lr){2-6}

        & \multirow{3}{*}{\shortstack[l]{
            \texttt{DeepSeek}\\
            $\rightarrow$ \texttt{Qwen3.5-4B}
        }}
        & All paired
        & 42.2\,{\scriptsize$(+29.3)$}
        & 23.3\,{\scriptsize$(+8.9)$}
        & 19.2\,{\scriptsize$(+9.8)$} \\
        & & Correctness-controlled
        & 34.9\,{\scriptsize$(+20.3)$}
        & 21.1\,{\scriptsize$(+8.0)$}
        & 14.4\,{\scriptsize$(+7.1)$} \\
        & & {\scriptsize \# tasks (controlled / all)}
        & {\scriptsize 36 / 150}
        & {\scriptsize 88 / 270}
        & {\scriptsize 189 / 221} \\
        \midrule

        \multirow{6}{*}{Mixed}
        & \multirow{3}{*}{\shortstack[l]{
            \texttt{Gemma}\\
            $\rightarrow$ \texttt{Qwen3.5-27B}
        }}
        & All paired
        & 60.0\,{\scriptsize$(-2.1)$}
        & 29.7\,{\scriptsize$(-22.1)$}
        & 13.9\,{\scriptsize$(-2.0)$} \\
        & & Correctness-controlled
        & 59.6\,{\scriptsize$(-2.2)$}
        & 29.4\,{\scriptsize$(-14.6)$}
        & 11.0\,{\scriptsize$(-3.4)$} \\
        & & {\scriptsize \# tasks (controlled / all)}
        & {\scriptsize 119 / 150}
        & {\scriptsize 116 / 270}
        & {\scriptsize 201 / 221} \\
        \cmidrule(lr){2-6}

        & \multirow{3}{*}{\shortstack[l]{
            \texttt{Qwen3.5-27B}\\
            $\rightarrow$ \texttt{Gemma}
        }}
        & All paired
        & 65.3\,{\scriptsize$(-0.5)$}
        & 39.2\,{\scriptsize$(+14.4)$}
        & 16.1\,{\scriptsize$(+2.0)$} \\
        & & Correctness-controlled
        & 66.0\,{\scriptsize$(+1.7)$}
        & 34.3\,{\scriptsize$(+12.5)$}
        & 13.7\,{\scriptsize$(+1.9)$} \\
        & & {\scriptsize \# tasks (controlled / all)}
        & {\scriptsize 119 / 150}
        & {\scriptsize 116 / 270}
        & {\scriptsize 201 / 221} \\
        \bottomrule
    \end{tabular}

    \caption{
        Post-swap performance under raw-history transfer on the agentic  environments, for all paired evaluations and the correctness-controlled
        subset.
        The controlled subset includes swap points where the source and recipient models have matching success/failure outcomes in every pre-swap session. Values in parentheses report $\mathrm{PS}_{\mathrm{swap}}-\mathrm{PS}_{\mathrm{self}}$, using the recipient's self score on the corresponding subset. The last row of each model pair reports the number of task sequences in the controlled subset and in all paired evaluations. \texttt{DeepSeek} and \texttt{Gemma} denote \texttt{DeepSeek-V4.1-Flash} and \texttt{Gemma-4-31B-it}, respectively.
    }
    \label{tab:memoryarena-swap-controlled}
\end{table*}

\section{Prompt}
\label{app:prompt}

We provide summarization prompts in \cref{fig:prompt-summary,fig:prompt-compaction}.

\begin{figure}[h]
    \centering
    \begin{paperprompt}{Session summarization prompt}
        \promptcontext{System prompt of the environment}\par
        \smallskip
        \promptcontext{One pre-swap session $j$: request $x_j$ and
        trajectory $\tau_j$, including reasoning}

        \tcblower
        \textbf{User:}
        Summarize the conversation above so that another assistant,
        who cannot see it, can continue the task from where it stopped.
    \end{paperprompt}

    \vspace{6pt}

    \begin{paperprompt}{Transferred context}
        \textbf{Hand-off summary of the earlier sessions}\par
        \smallskip
        This is a summary of the earlier sessions of this task.
        For each session, the user's request is given verbatim,
        followed by a summary of what the previous assistant did
        in that session.

        \tcblower
        \handofflabel{=== Session 1: the user's request ===}
        $x_1$ \textit{(verbatim)}

        \smallskip
        \handofflabel{=== Session 1:\\
        summary of the previous assistant's work ===}
        $s(\tau_1)$

        \par\vspace{3pt}
        {\centering $\vdots$\par}
        \vspace{3pt}

        \handofflabel{=== Session $i$: the user's request ===}
        $x_i$ \textit{(verbatim)}

        \smallskip
        \handofflabel{=== Session $i$:\\
        summary of the previous assistant's work ===}
        $s(\tau_i)$
    \end{paperprompt}
    \vspace{.5 em}
    \caption{
        Session-wise summarization and the transferred context for a single swap after session $i$.
        Each session is summarized independently, with the original request retained verbatim.
    }
    \label{fig:prompt-summary}
\end{figure}

\begin{figure}[h]
    \centering
    \begin{paperprompt}{Web Search context compaction prompt}
        \promptcontext{System prompt of the environment}\par
        \smallskip
        \promptcontext{The earlier part of the context to be compacted}

        \tcblower
        \textbf{User:}
        Summarize the earlier part of this conversation so that you
        can continue the task without it. Keep every fact found,
        entity, name, date and number, every constraint of the task,
        every answer given and how it was reached, and the
        sub-questions that are still open. Drop redundant or
        irrelevant search results and repeated queries. Write the
        summary as plain notes; do not answer the task and do not
        call tools.
    \end{paperprompt}
    \vspace{.5 em}
    \caption{
        Prompt for context compaction in Web Search.
    }
    \label{fig:prompt-compaction}
\end{figure}

\section{Use of LLM Assistants}
We used LLM assistants to support experiments and analysis and to improve the clarity and readability of the manuscript. The authors take full responsibility for the experimental results, their interpretation, and the final content of the manuscript.



\end{document}